\documentclass[10pt,journal,cspaper,compsoc]{IEEEtran}
\usepackage{amsfonts}
\ifCLASSOPTIONcompsoc
\usepackage[nocompress]{cite}
\else
\usepackage{cite}
\fi
\usepackage{amsmath}
\usepackage{algorithm}
\usepackage{algorithmic}
\usepackage{stfloats}
\usepackage{url}
\usepackage{mathrsfs}
\usepackage{dsfont}
\usepackage{multirow}
\usepackage{multicol}
\usepackage{color}
\usepackage[colorlinks,linkcolor=red]{hyperref}
\usepackage{graphics}
\usepackage{graphicx}
\usepackage{subfig}
\usepackage{xcolor}
\usepackage[normalem]{ulem} 
\usepackage{booktabs}
\newcommand\hl{\bgroup\markoverwith
	{\textcolor{yellow}{\rule[-.5ex]{2pt}{2.5ex}}}\ULon}

\begin{document}
	
	\title{Pure Vision-Language-Action (VLA) Models: \\ A Comprehensive Survey}

\author{Dapeng Zhang*\textsuperscript{\rm 1,2}, Jing Sun*\textsuperscript{\rm 1}, Chenghui Hu*\textsuperscript{\rm 1}, Xiaoyan Wu\textsuperscript{\rm 1}, Zhenlong Yuan\textsuperscript{\rm 3}, Rui Zhou\textsuperscript{\rm 1}, \\ Fei Shen\textsuperscript{\dag}\textsuperscript{\rm 2}, and Qingguo Zhou\textsuperscript{\dag}\textsuperscript{\rm 1}




\IEEEcompsocitemizethanks{
\IEEEcompsocthanksitem{
    \textsuperscript{\rm 1} Lanzhou University, China\\
    \textsuperscript{\rm 2} National University of Singapore, Singapore\\
    \textsuperscript{\rm 3}Institute of Computing Technology, Chinese Academy of Sciences, China\\
    }

 \IEEEcompsocthanksitem * Equal contributions
 \IEEEcompsocthanksitem \textsuperscript{\dag} Fei Shen and Qingguo Zhou are the corresponding authors 
\IEEEcompsocthanksitem D. Zhang, J. Sun, C. Hu, X. Wu, R. Zhou and Q. Zhou are with School of Information Science and Engineering, Lanzhou University, China. E-mail: \{zhangdp22, sjing2023,huchh2024,wuxiaoyan2024,zr,zhouqg\}@lzu.edu.cn

\IEEEcompsocthanksitem Z. Yuan is with the Institute of Computing Technology, Chinese Academy of Sciences, China. E-mail: yuanzhenlong21b@ict.ac.cn 

\IEEEcompsocthanksitem  D. Zhang and F. Shen are with the NExT++ Research Centre, National University of Singapore, Singapore. E-mail: shenfei29@nus.edu.sg

}
}
	
\IEEEcompsoctitleabstractindextext{
\begin{abstract}
The emergence of Vision-Language-Action (VLA) models marks a paradigm shift from traditional policy-based control to generalized robotics, reframing Vision-Language Models (VLMs) from passive sequence generators into active agents for manipulation and decision-making in complex, dynamic environments. This survey delves into advanced VLA methods, aiming to provide a clear taxonomy and a systematic, comprehensive review of existing research. It presents a comprehensive analysis of VLA applications across different scenarios and classifies VLA approaches into several paradigms: autoregression-based, diffusion-based, reinforcement-based, hybrid, and specialized methods; while examining their motivations, core strategies, and implementations in detail. In addition, foundational datasets, benchmarks, and simulation platforms are introduced. Building on the current VLA landscape, the review further proposes perspectives on key challenges and future directions to advance research in VLA models and generalizable robotics. By synthesizing insights from over three hundred recent studies, this survey maps the contours of this rapidly evolving field and highlights the opportunities and challenges that will shape the development of scalable, general-purpose VLA methods.
\end{abstract}
		
}

\maketitle

\IEEEdisplaynotcompsoctitleabstractindextext
\IEEEpeerreviewmaketitle


\section{Introductions} \label{sec:intro}

Robotics has long been a prominent area of scientific research. Historically, robots primarily relied on pre-programmed instructions and engineered control policies to perform task decomposition and execution. These methods were commonly applied to simple, repetitive tasks, such as factory assembly lines and logistics sorting. In recent years, the rapid advancement of artificial intelligence has enabled researchers to exploit the feature extraction and trajectory prediction capabilities of deep learning across diverse modalities, including images, text, and point clouds. By integrating techniques such as perception, detection, tracking, and localization, researchers have decomposed robotic tasks into multiple stages to meet execution requirements, thereby advancing the development of embodied intelligence and autonomous driving. However, most of these robots still operate as isolated agents, designed for specific tasks and lacking effective interaction with humans and external environment.

\begin{figure}
    \centering
    \includegraphics[width=0.9\linewidth]{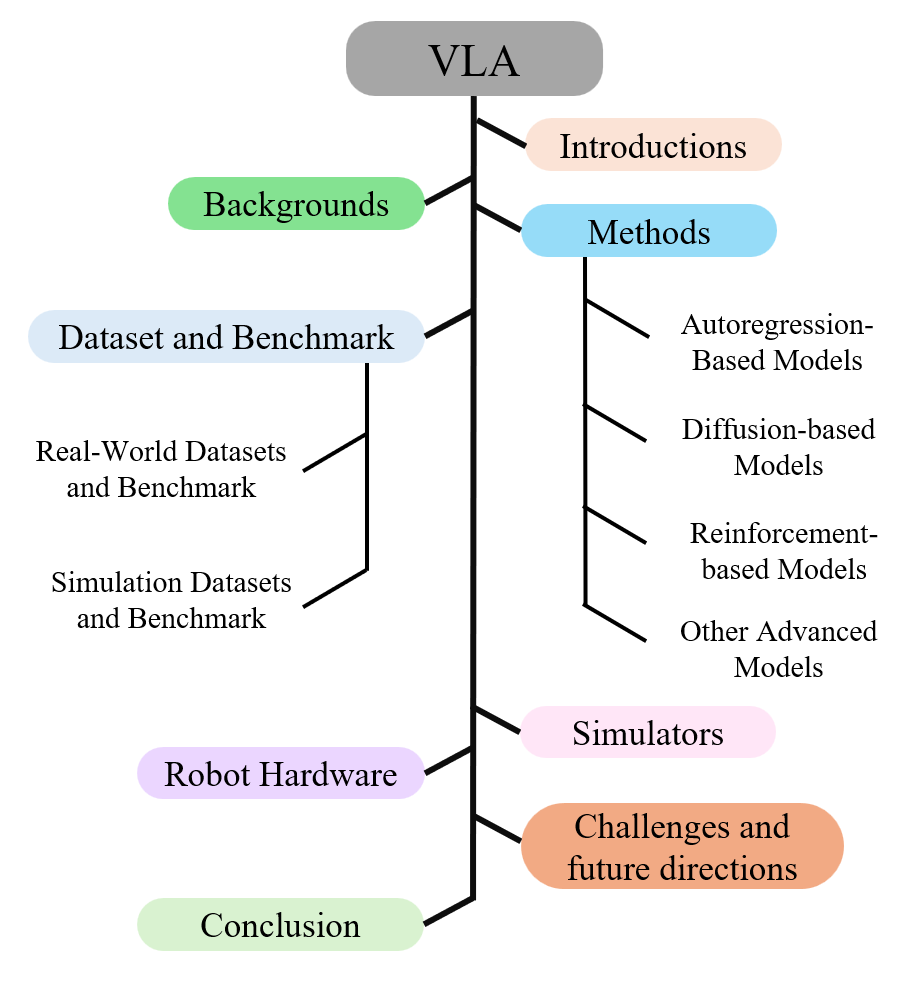}
    \vspace{-0.3cm}
    \caption{Organization and Structure of the VLA Survey.}
    \vspace{-0.7cm}
    \label{fig:structure}
\end{figure}

To address these limitations, researchers have begun exploring the incorporation of large language models (LLMs) and vision-language models (VLMs) to enable more accurate and flexible robotic manipulation. Modern robotic manipulation methods \cite{wen2025diffusionvlageneralizableinterpretablerobot, black2024pi0visionlanguageactionflowmodel} typically leverage vision language generative paradigms (e.g., autoregressive models \cite{liu2024clipsenhancedclipframework,li2022blipbootstrappinglanguageimagepretraining,xue2025xgenmmblip3familyopen,bai2025qwen25vltechnicalreport} or diffusion models \cite{chi2024diffusionpolicyvisuomotorpolicy}), combined with large-scale datasets \cite{o2024open} and advanced fine-tuning strategies. We refer to these as VLA foundation models, which have substantially improved the quality of robotic manipulations. Fine-grained action control over generated content provides users with greater flexibility, unlocking the practical potential of VLA for task execution.

Despite their promise, reviews of pure VLA methods remain scarce. Existing surveys either focus on taxonomy over VLM foundational models or provide broad overviews of robotic manipulation as a whole. Firstly, VLA methods represent a nascent field in robotics, with no established methodological landscape or consensus taxonomy, making it challenging to systematically summarize these approaches. Secondly, current reviews either classify VLA approaches based on differences in foundational models or present a comprehensive analysis of robotic applications across the entire history of the field, often emphasizing traditional methods at the expense of emerging techniques. While these reviews offer valuable insights, they provide only cursory examinations of robotic models or concentrate primarily on foundational models, leaving a significant gap in the literature regarding pure VLA methods.

\begin{figure}
    \centering
    \includegraphics[width=0.9\linewidth]{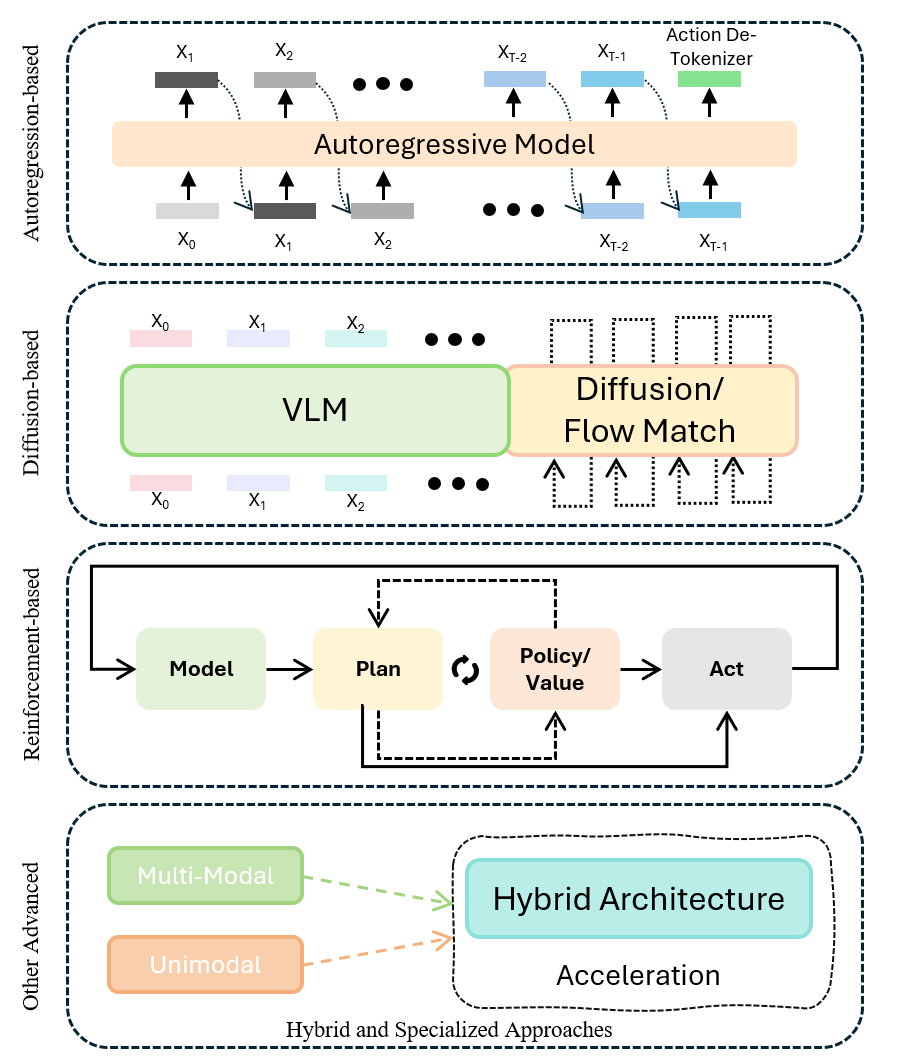}
    \vspace{-0.3cm}
    \caption{ Illustration of various VLA skeleton.}
    \vspace{-0.7cm}
    \label{fig:vla_3}
\end{figure}

In this paper, we investigate VLA methods and associated resources, providing a focused and comprehensive review of existing approaches. Our goal is to present a clear taxonomy, systematically summarize VLA research, and elucidate the development trajectory of this rapidly evolving field. After a brief overview of LLMs and VLMs, we focus on the policy strategies of VLA models, highlighting the unique contributions and distinctive features of previous studies. We classify VLA approaches into 4 categories: autoregression-based, diffusion-based, reinforcement-based, hybrid, and specialized methods, and provide a detailed analysis of their motivations, core strategies, and mechanisms. As shown in Fig. \ref{fig:vla_3}, we present a VLA skeleton of these methods. 
We examine application domains, including robotic arms, quadruped robots, humanoids, and wheeled robots (autonomous vehicles), offering a comprehensive assessment of VLA deployment across diverse scenarios. Given the strong dependence of VLA models on datasets and simulation platforms, we provide a concise overview of these resources. Finally, based on the current VLA landscape, we identify key challenges and outline future research directions—including data limitations, inference speed, and safety—to accelerate the advancement of VLA models and generalizable robotics.

The overall structure of this survey is illustrated in Fig. \ref{fig:structure}. First, Section \ref{sec:back} provides an overview of the background for VLA research. Section \ref{sec:method} presents the existing VLA approaches in the robotics field. Section \ref{sec:data} introduces the datasets and benchmarks employed by VLA approaches. Sections \ref{sec:sim} and \ref{sec:hardw} discuss simulation platforms and robotic hardware. Section \ref{sec:chall} further discusses the challenges and future directions for VLA-based robotic methods. Finally, we summarize the paper and provide our perspective on future developments.

In summary, our contributions are as follows:
\begin{itemize}

\item We present the well-structured taxonomy of pure VLA methods, classifying approaches based on their action-generation strategies. This facilitates understanding of existing methods and highlights core challenges in the field.

\item The survey emphasizes the defining characteristics and methodological innovations of each category and technique, providing a clear perspective on current approaches.

\item We provide a comprehensive overview of associated resources (datasets, benchmarks and simulation platforms) for training and evaluating VLA models.

\item We investigate the practical impact of VLA in robotics, identify key limitations of existing techniques, and propose potential avenues for further exploration.

\end{itemize}

\begin{figure*}
    \centering
    \includegraphics[width=0.9\linewidth]{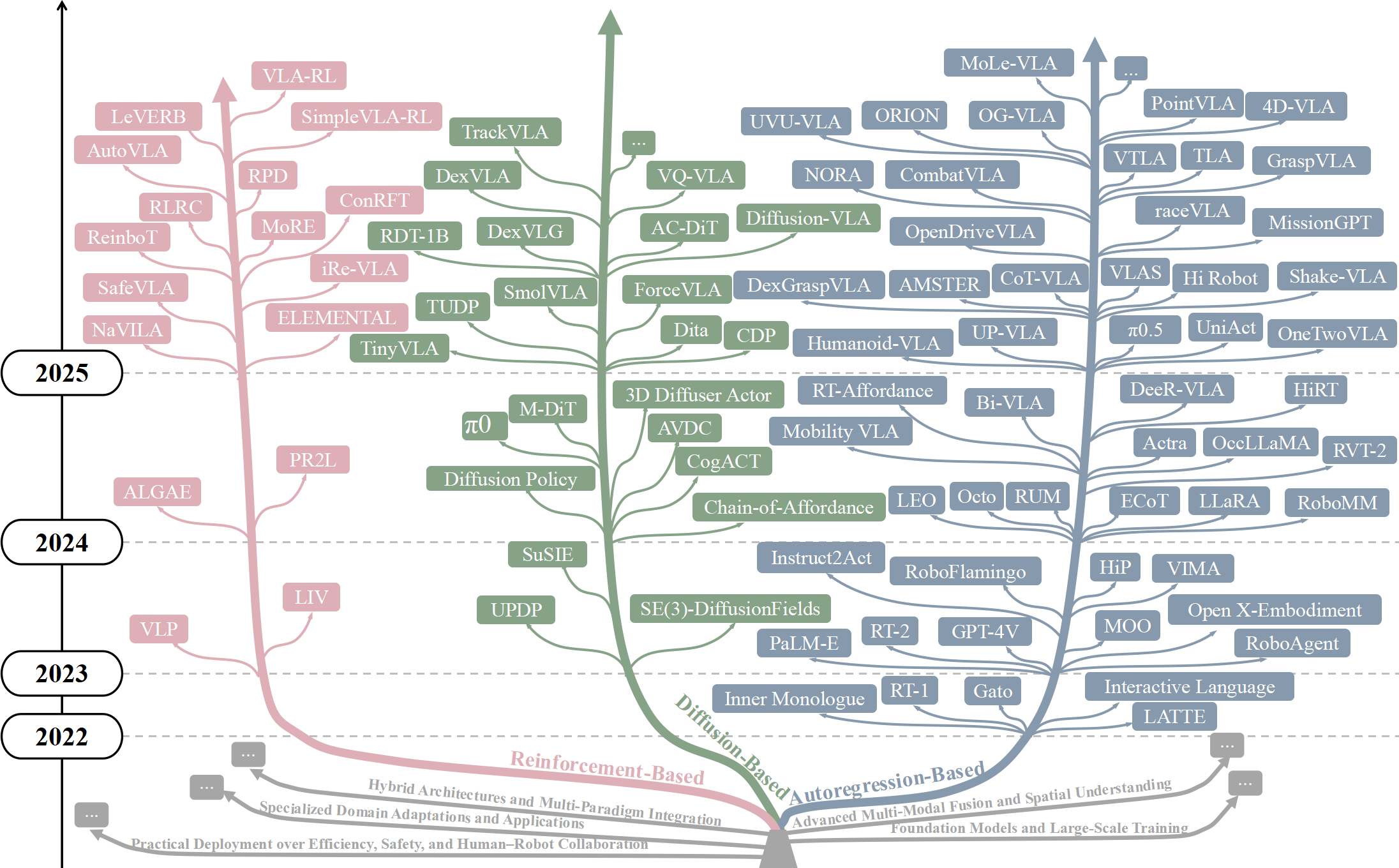}
    \vspace{-0.3cm}
    \caption{Vision-Language-Action Taxonomy: From Autoregression-based, Diffusion-based, to Reinforcement-based and Hybrid/Specialized methods, multi-paradigm advances and practical adaptations in VLA. The taxonomy is organized along a chronological timeline.}
    \vspace{-0.6cm}
    \label{fig:timeline}
\end{figure*}

\section{Backgrounds}\label{sec:back}

The emergence of Vision-Language-Action (VLA) models represents a significant advance toward general-purpose embodied intelligence. Traditional robotic systems typically rely on isolated perception pipelines, hand-engineered control strategies, or task-specific reinforcement learning. Although these approaches perform effectively in constrained environments, such as factory floors or laboratories, they generalize poorly to dynamic and unstructured settings. Modern robots can “see” using computer vision models, “understand” language through large language models, and “act” via controllers or learned policies; however, integrating these capabilities into a coherent and unified system remains a key challenge. VLA models address this challenge by providing a unified framework that grounds language in perception and maps it to executable actions.

\subsection{Early: LLM/VLM fundamental Models}

Breakthroughs in unimodal modeling have laid the methodological and engineering foundation for multimodal integration. In computer vision, convolutional neural networks (e.g., AlexNet \cite{krizhevsky2012imagenet}, ResNet \cite{he2016deep}) established a representational paradigm transitioning from local convolutions to deep residual learning, which was further advanced by the Vision Transformer (ViT) \cite{dosovitskiy2020image}. ViT introduced self-attention to the image domain, markedly improving transferability and generalization. In natural language processing, the Transformer architecture enabled large-scale pretraining and alignment techniques, giving rise to models such as BERT \cite{devlin2019bert}, GPT \cite{radford2018improving}, T5 \cite{raffel2020exploring}, and GPT-4 \cite{achiam2023gpt}, which exhibit strong reasoning capabilities, instruction-following, and in-context learning. In parallel, reinforcement learning advanced policy optimization and sequential decision-making, progressing from DQN and PPO to Decision Transformer, which highlights a unifying perspective of control through sequence modeling.

Against this backdrop, vision-language models (VLMs) emerged as a crucial bridge between unimodal learning and embodied intelligence. Early approaches (e.g., ViLBERT \cite{lu2019vilbert}, VisualBERT \cite{li2019visualbert}) aligned and fused images and text using dual-stream or single-stream Transformers, while contrastive learning methods (e.g., CLIP) mapped large-scale image–text pairs into a shared embedding space, enabling zero-shot and few-shot recognition and retrieval. More recently, instruction-tuned, dialogue-centric multimodal models (e.g., BLIP-2 \cite{li2023blip}, Flamingo \cite{alayrac2022flamingo}, LLaVA \cite{liu2023visual}) have substantially strengthened open-ended cross-modal understanding, fine-grained grounding, and multi-turn reasoning, setting the stage for Vision-Language-Action (VLA) systems.

\subsection{Present: Development of VLA Models}
\subsubsection{From LLM/VLM to VLA Models}

Following this trajectory, research has naturally advanced toward VLA integration, which unifies visual perception, language understanding, and executable control within a single sequence modeling framework\cite{kim2024openvlaopensourcevisionlanguageactionmodel, pertsch2025fastefficientactiontokenization,wen2025diffusionvlageneralizableinterpretablerobot, black2024pi0visionlanguageactionflowmodel,yang2025instructvla}. Typical designs encode images and instructions as prefix or context tokens, inject robot states and sensory feedback as state tokens, and autoregressively generate action tokens to produce control sequences, thereby closing the perception-language-action loop. Compared with traditional perception, planning, and control pipelines, VLA provides end-to-end cross-modal alignment and a unified treatment of goals, constraints, and intent. It inherits the semantic and instruction generalization of VLMs, while explicit state coupling and action generation confer robustness to environmental disturbances and long-horizon tasks. This progression—from unimodal to multimodal, and then to multimodal plus executable control—establishes the methodological foundation for systems that can not only see and understand, but also act.

\subsubsection{The Supporting Role of Data and Simulation}

The development of Vision-Language-Action (VLA) models in robotics relies heavily on high-quality datasets and realistic simulators capable of capturing the complexity of real-world scenarios. Recent robotic methods are typically deep learning–based and data-driven; consequently, dataset collection and annotation play a critical role in driving advancements in the field. Some datasets are collected in real-world settings, which requires substantial human effort and financial resources. To address these challenges, researchers also leverage large-scale human manipulation videos from the Internet as generalization datasets to provide auxiliary supervision for VLA model training. Despite these efforts, data collection remains costly, annotations are labor-intensive, and long-tail corner cases are often underrepresented. 
Other datasets are generated through robot simulators, which facilitate the collection of large-scale labeled data. Simulators provide diverse and controllable environments, flexible sensor configurations, realistic kinematic models, and interactive static and dynamic scenes, supporting both data collection and model evaluation. Representative datasets include Open X-Embodiment (OXE) \cite{o2024open}, which integrates 22 robot datasets from 21 institutions, encompassing 527 skills and 160,266 tasks, and BridgeData \cite{ebert2021bridge}, comprising 71 tasks across 10 environments in multiple domains. These resources standardize data formats, thereby promoting rapid development and reproducibility in VLA research. Simulators such as THOR \cite{kolve2017ai2}, Habitat \cite{savva2019habitat}, MuJoCo \cite{todorov2012mujoco}, Isaac Gym \cite{makoviychuk2021isaacgymhighperformance}, and CARLA \cite{carla} offer extensible virtual environments capable of generating multimodal annotations, including action trajectories, object states, and natural language instructions. Collectively, these datasets and simulation platforms mitigate the scarcity of real-world robot data and accelerate the training and evaluation of VLA models.

\subsection{Future: Towards General Embodied Intelligence}

VLA models occupy the forefront of research where vision, language, and action converge. They build upon breakthroughs in foundational models for perception and reasoning, emphasizing capabilities in human interaction and task execution, and extend these abilities to the physical world. By integrating the representational power of visual encoders, the reasoning capabilities of large language models, and the decision-making abilities of reinforcement learning and control frameworks, VLA models hold significant potential to bridge the “perception–understanding–action” gap. Despite facing challenges related to scalability, generalization, safety, and real-world deployment, VLA is widely recognized as a key frontier in embodied artificial intelligence. Although VLA has achieved notable success in Vision-Language-Action interactions, and benefits from advances in large-scale language models, it has not yet attained full generality in the domain of embodied intelligence.
General embodied intelligence posits that human-like intelligent behavior relies not only on cognitive processing but also on the physical body, environmental perception, and feedback mechanisms, enabling interaction with the external world. To accommodate the demands of diverse tasks, general embodied intelligence can manifest through various types of robots, including humanoid robots for household applications, assembly robots with dexterous manipulators, and bionic robots with specialized capabilities. Clearly, general embodied intelligence has the potential to enable artificial systems to perform a wider array of tasks across diverse environments. VLA is currently evolving toward this vision of general embodied intelligence and holds considerable promise for realizing it.

\section{Vision-Language-Action Models}\label{sec:method}

In recent years, Vision-Language-Action (VLA) models have undergone rapid and systematic development, driven by advances in multimodal representation learning, generative modeling, and reinforcement learning. To trace this evolution, this section reviews the major methodological paradigms in VLA, including autoregressive-based modeling, diffusion-based approaches, reinforcement learning strategies, and hybrid or specialized designs. Fig.~\ref{fig:timeline} presents a tree diagram illustrating the progression of these paradigms, with each branch highlighting representative works within the respective taxonomy. The taxonomy is organized chronologically, emphasizing how methodological innovations have progressively expanded the capabilities of VLA models.

\subsection{Autoregression-Based Models in Vision-Language-Action Research}

Autoregression-based models constitute a classical yet effective paradigm for sequence generation in Vision-Language-Action (VLA) tasks. By treating action sequences as temporally dependent processes, these models generate actions step by step, conditioned on prior context, perceptual inputs, and task prompts. With the rapid advancement of Transformer architectures, recent VLA systems~\cite{brohan2023rt1roboticstransformerrealworld,kim2024openvlaopensourcevisionlanguageactionmodel,cen2025worldvla} have demonstrated the scalability and robustness of this approach. 
Representative works in these directions are summarized in Table~\ref{tab:merged-vla}, collectively highlighting the versatility and generality of the autoregression-based modeling paradigm in VLA research.

\begin{table*}[!htpb]
\centering
\caption{Representative works on Autoregression-Based Models.}
\label{tab:merged-vla}
\footnotesize
\begin{tabular}{p{3.5cm}p{1cm}p{12.5cm}}
\toprule
\textbf{Method / Model} & \textbf{Year} & \textbf{Key Innovation in Autoregression-Based Models} \\
\midrule

\multicolumn{3}{l}{\textbf{(A) Autoregressive Generalist VLA Methodologies}} \\
\midrule
Gato~\cite{reed2022generalistagent} & 2022 & Unified multimodal Transformer with tokenized vision, language, states, and actions. \\
RT-1~\cite{brohan2023rt1roboticstransformerrealworld} & 2022 & Real-world robotic Transformer trained on 130k demos with FiLM-based multimodal fusion. \\
PaLM-E~\cite{driess2023palmeembodiedmultimodallanguage} & 2023 & Embodied multimodal LLM combining PaLM and ViT for VQA, navigation, and manipulation. \\
RT-2~\cite{brohan2023rt2visionlanguageactionmodelstransfer} & 2023 & Extended PaLM-E with action tokens and web-scale VLM knowledge for open-vocabulary grasping. \\
LEO~\cite{huang2024embodiedgeneralistagent3d} & 2024 & Two-stage training for 3D vision–language alignment and VLA fine-tuning. \\
Octo~\cite{octomodelteam2024octoopensourcegeneralistrobot} & 2024 & Open-source cross-robot policy trained on 1.5M video instances with reward-free imitation. \\
RUM~\cite{etukuru2024robotutilitymodelsgeneral} & 2024 & Utility-based scoring enabling robust zero-shot deployment and new benchmarks. \\
RoboMM~\cite{yan2024robommallinonemultimodallarge} & 2024 & Multimodal fusion with modality masking achieving SOTA on RoboData. \\
UP-VLA~\cite{zhang2025upvlaunifiedunderstandingprediction} & 2025 & Unified prompting framework merging tasks, visuals, and actions for better few-shot performance. \\
UniAct~\cite{zheng2025universalactionsenhancedembodied} & 2025 & Defined universal atomic actions to solve cross-embodiment heterogeneity. \\
Humanoid-VLA~\cite{ding2025humanoidvlauniversalhumanoidcontrol} & 2025 & Applied VLA to humanoid control with multi-view RGB and language prompts. \\
NORA~\cite{hung2025norasmallopensourcedgeneralist} & 2025 & Lightweight open-source VLA using FAST+ tokenizer and 970k demonstrations. \\
OneTwoVLA~\cite{lin2025onetwovlaunifiedvisionlanguageactionmodel} & 2025 & Adaptive System 1 \& 2 reasoning for long-horizon planning and error recovery. \\
VOTE~\cite{lin2025votevisionlanguageactionoptimizationtrajectory} & 2025 & Voting-based ensemble reduces action tokens for faster inference and training. \\
UniVLA~\cite{bu2025univlalearningacttaskcentric} & 2025 & Learned task-centric latent action representations from diverse videos. \\
OE-VLA~\cite{zhao2025unveilingpotentialvisionlanguageactionmodels} & 2025 & Extended VLAs for open-ended multimodal instruction following. \\
\midrule
\multicolumn{3}{l}{\textbf{(B) Autoregressive Reasoning and Semantic Planning with LLMs}} \\
\midrule
Inner Monologue~\cite{huang2022innermonologueembodiedreasoning} & 2022 & Introduced inner language-driven reasoning loop, improving embodied task success. \\
Instruct2Act~\cite{huang2023instruct2actmappingmultimodalityinstructions} & 2023 & Proposed vision–language–task script–action pipeline with semantic intermediaries. \\

RoboFlamingo~\cite{li2024visionlanguage} & 2023 & Adapted OpenFlamingo to robotics, enabling efficient VLM-to-VLA transfer. \\

LLaRA~\cite{li2024llara} & 2024 & Augmented trajectories with dialogue tasks to enhance data and transfer. \\
ECoT~\cite{zawalski2025roboticcontrolembodiedchainofthought} & 2024 & Formalized embodied chain-of-thought reasoning to boost OpenVLA. \\
Mobility VLA~\cite{xu2024mobility} & 2024 & Integrated long-context VLM with navigation for multimodal instruction-following. \\

UAV-VLA~\cite{10.5555/3721488.3721725} & 2025 & Generated UAV missions from satellite imagery using VLM-based planning. \\
Tactile-VLA~\cite{huang2025tactilevlaunlockingvisionlanguageactionmodels} & 2025 & Combined tactile feedback with reasoning for precise, adaptive control. \\
Shake-VLA~\cite{khan2025shakevlavisionlanguageactionmodelbasedbimanual} & 2025 & Introduced multimodal bimanual system with speech, vision, and haptics. \\

Hi Robot~\cite{shi2025hirobotopenendedinstruction} & 2025 & Adopted hierarchical coarse-to-fine planning for long-instruction following. \\
DexGraspVLA~\cite{zhong2025dexgraspvlavisionlanguageactionframeworkgeneral} & 2025 & Combined VLM planning with diffusion for robust dexterous grasping. \\
HAMSTER~\cite{li2025hamsterhierarchicalactionmodels} & 2025 & Hierarchical VLA leveraging out-of-domain data for broad generalization. \\

CoT-VLA~\cite{zhao2025cotvlavisualchainofthoughtreasoning} & 2025 & Implemented visual chain-of-thought reasoning with predictive goals. \\
Gemini Robotics~\cite{geminiroboticsteam2025geminiroboticsbringingai} & 2025 & Built Gemini 2.0-based platform for robust multi-task embodied reasoning. \\

CognitiveDrone~\cite{lykov2025cognitivedrone} & 2025 & Developed cognitive UAV VLA achieving high success on aerial benchmark. \\
$\pi$0.5~\cite{intelligence2025pi05visionlanguageactionmodelopenworld} & 2025 & Trained on heterogeneous robots for open-world generalization. \\

InSpire~\cite{zhang2025inspire} & 2025 & Introduced spatial reasoning prompts to reduce spurious correlations. \\

\midrule
\multicolumn{3}{l}{\textbf{(C) Autoregressive Trajectory Generation and Visual Alignment Modeling}} \\
\midrule
LATTE~\cite{bucker2022lattelanguagetrajectorytransformer} & 2022 & Mapped natural language to motion trajectories with a Transformer decoder. \\
VIMA~\cite{jiang2023vimageneralrobotmanipulation} & 2023 & Unified multimodal tokens for language, vision, and actions with strong zero-shot generalization. \\
InstructRL~\cite{liu2023instructionfollowingagentsmultimodaltransformer} & 2023 & Used joint vision–language encoders with policy Transformer for better alignment. \\

CrossFormer~\cite{doshi2024scalingcrossembodiedlearningpolicy} & 2024 & Scalable Transformer policy handling multi-embodiment data for generalization. \\
GR-1~\cite{wu2024unleashing} & 2024 & Unified video and action tokens for temporal modeling via video pretraining transfer. \\
GR-2~\cite{cheang2024gr2generativevideolanguageactionmodel} & 2024 & Extended GR-1 with web-scale video–language supervision for zero-shot manipulation. \\
OpenVLA~\cite{kim2024openvlaopensourcevisionlanguageactionmodel} & 2024 & Released 7B open-source VLA trained on 970k trajectories, surpassing RT-2-X. \\
Bi-VLA~\cite{gbagbe2024bivlavisionlanguageactionmodelbasedbimanual} & 2024 & Employed dual-arm predictors for coordinated bimanual manipulation. \\

RoboNurse-VLA~\cite{li2024robonursevlaroboticscrubnurse} & 2024 & Achieved high-precision surgical grasping with VLA policies in medical settings. \\

Moto~\cite{chen2025motolatentmotiontoken} & 2025 & Bridged video pretraining and execution with motion language tokens. \\

OpenDriveVLA~\cite{zhou2025opendrivevlaendtoendautonomousdriving} & 2025 & Aligned 2D/3D perception into a unified semantic space for driving trajectories. \\

GraspVLA~\cite{deng2025graspvlagraspingfoundationmodel} & 2025 & Pretrained GPT-style decoder with GraspVerse for real-world grasp transfer. \\

RaceVLA~\cite{serpiva2025racevlavlabasedracingdrone} & 2025 & Generated vision–language trajectories for dynamic drone racing control. \\
VTLA~\cite{zhang2025vtlavisiontactilelanguageactionmodelpreference} & 2025 & Fused vision–tactile input with preference optimization for \textgreater 90\% unseen success. \\

PointVLA~\cite{li2025pointvlainjecting3dworld} & 2025 & Injected point clouds into pretrained VLAs for lightweight 3D reasoning. \\

WorldVLA~\cite{cen2025worldvla} & 2025 & Mitigated autoregressive error propagation with joint vision–action modeling. \\
TraceVLA~\cite{zheng2025tracevlavisualtraceprompting} & 2025 & Used visual trace prompting to capture cues in long-horizon tasks. \\

Interleave-VLA~\cite{fan2025interleavevlaenhancingrobotmanipulation} & 2025 & Supported interleaved image–text inputs to boost zero-shot manipulation. \\
MoManipVLA~\cite{Wu_2025_CVPR} & 2025 & Unified base–arm trajectory alignment for mobile manipulation transfer. \\
\midrule
\multicolumn{3}{l}{\textbf{(D) Structural Optimization and Efficient Inference Mechanisms in Autoregressive VLA}} \\
\midrule

HiP~\cite{ajay2023compositionalfoundationmodelshierarchical} & 2023 & Introduced hierarchical planning with token prediction for long-horizon manipulation. \\

Actra~\cite{ma2024actraoptimizedtransformerarchitecture} & 2024 & Optimized Transformer with trajectory attention and learnable queries to cut overhead. \\

DeeR-VLA~\cite{NEURIPS2024_67b0e7c7} & 2024 & Multi-exit architecture enabling early exits to reduce latency. \\
FAST~\cite{pertsch2025fastefficientactiontokenization} & 2025 & Generated variable-length action tokens for efficient long-horizon execution. \\
SpatialVLA~\cite{qu2025spatialvlaexploringspatialrepresentations} & 2025 & Enhanced geometric perception using voxel grids and spatial attention. \\
VLA-Cache~\cite{xu2025vlacacheefficientvisionlanguageactionmodel} & 2025 & Reused Transformer key–value states with adaptive caching for efficiency. \\
Beyond Sight~\cite{jones2025sightfinetuninggeneralistrobot} & 2025 & Language-guided perception adaptation improved multimodal sensor fusion. \\
MoLe-VLA~\cite{zhang2025molevladynamiclayerskippingvision} & 2025 & Mixture-of-experts router enabled dynamic layer skipping, reducing cost ~40\%. \\
PD-VLA~\cite{song2025acceleratingvisionlanguageactionmodelintegrated} & 2025 & Applied parallel fixed-point decoding to accelerate inference without retraining. \\
BitVLA~\cite{wang2025bitvla1bitvisionlanguageactionmodels} & 2025 & Used 1-bit quantization, cutting memory to 30\% while preserving performance. \\
GR-MG~\cite{li2025gr} & 2025 & Leveraged diffusion-generated goals for semi-supervised autoregressive learning. \\
LoHoVLA~\cite{yang2025lohovlaunifiedvisionlanguageactionmodel} & 2025 & Unified hierarchical control for ultra-long-horizon closed-loop tasks. \\
OTTER~\cite{huang2025ottervisionlanguageactionmodeltextaware} & 2025 & Injected language into vision encoding with TAVE for stronger alignment. \\
ChatVLA~\cite{zhou2025chatvlaunifiedmultimodalunderstanding} & 2025 & Unified architecture with expert routing and phased alignment for scalability. \\

\bottomrule
\end{tabular}
\end{table*}

\subsubsection{Autoregressive Generalist VLA Methodologies}


Research on generalist VLA agents unifies perception, task instructions, and action generation within autoregressive sequence modeling. By tokenizing multimodal inputs, these models enable step-by-step action generation across heterogeneous tasks.


Early work like Gato \cite{reed2022generalistagent} demonstrated tokenizing heterogeneous modalities for joint training. Subsequent scaling efforts, notably RT-1/RT-2 \cite{brohan2023rt1roboticstransformerrealworld,brohan2023rt2visionlanguageactionmodelstransfer} leveraged massive real-world datasets and web-scale pretraining, while PaLM-E \cite{driess2023palmeembodiedmultimodallanguage} integrated pretrained language knowledge into embodied control, establishing autoregressive Transformers as practical unifying models.


To address embodiment fragmentation, frameworks like Octo \cite{octomodelteam2024octoopensourcegeneralistrobot}, LEO \cite{huang2024embodiedgeneralistagent3d}, and UniAct \cite{zheng2025universalactionsenhancedembodied} aligned visual-linguistic modalities with universal action abstractions for cross-platform compatibility.
Recent advances focus on reasoning integration and efficiency. Models now combine action generation with language reasoning and adaptive prompting for long-horizon planning \cite{kim2024openvlaopensourcevisionlanguageactionmodel,zhang2025upvlaunifiedunderstandingprediction, lin2025onetwovlaunifiedvisionlanguageactionmodel}. Lightweight designs like NORA \cite{hung2025norasmallopensourcedgeneralist} and RoboMM \cite{yan2024robommallinonemultimodallarge} address deployment constraints.


Overall, research on generalist VLA agents has progressed from early unified tokenization to large-scale real-world training and semantic grounding, advancing toward cross-platform universality, reasoning integration, and efficiency-oriented designs. This trajectory reflects a shift from proof-of-concept demonstrations to systems emphasizing scalability, semantic reasoning, and deployability. Table~\ref{tab:merged-vla} (A) summarizes representative autoregressive generalist agents and their key contributions. However, issues of safety, interpretability, and alignment with human values remain largely unresolved, leaving ample room for future research.

\subsubsection{Autoregressive Reasoning and Semantic Planning with LLMs} 


The integration of LLMs has transformed them from passive input parsers into semantic mediators within VLA systems, enabling reasoning-driven control for long-horizon and compositional tasks. This section reviews LLM-based reasoning evolution from semantic intermediaries to hierarchical planners and platform-level orchestration.


To inject reasoning into VLA models, Inner Monologue \cite{huang2022innermonologueembodiedreasoning,huang2023instruct2actmappingmultimodalityinstructions} introduced self-talk-style reasoning with pre-action planning and post-action reflection. Extensions like Prompt-to-Walk, RoboFlamingo, and RoboMM \cite{wang2024promptrobotwalklarge,li2024visionlanguage,yan2024robommallinonemultimodallarge} demonstrated linguistic representations across locomotion and manipulation tasks.
 

Subsequent approaches enhanced adaptability through feedback and hierarchical planning. Interactive Language \cite{lynch2022interactivelanguagetalkingrobots} enabled real-time correction, Open-Ended Instructable Agents \cite{sarch2023openendedinstructableembodiedagents} utilized episodic memory, and Hi Robot \cite{shi2025hirobotopenendedinstruction} adopted hierarchical planning for long instructions. MissionGPT, Mobility VLA, and NORA \cite{10852423,xu2024mobility,hung2025norasmallopensourcedgeneralist} emphasized lightweight deployment and dialogue-driven adaptability.


Hierarchical frameworks combined semantic planning with controllers for dexterous manipulation \cite{bharadhwaj2023roboagent, hu2023lookleapunveilingpower, 10865585,nasiriany2024rtaffordanceaffordancesversatileintermediate}. InSpire, From Foresight to Forethought, and CoT-VLA \cite{zhang2025inspire,wu2025foresightforethoughtvlminthelooppolicy,zhao2025cotvlavisualchainofthoughtreasoning} emphasized runtime stability and chain-of-thought mechanisms.

Autoregression-based reasoning architecture commonly patches the input to the sequence and utilizes those tokens for further reasoning. These models can process input of various lengths,and the strong ability of in-context learning enables them to process different modalities under aunified structure \cite{zhao2025vlasvisionlanguageactionmodelspeech}. UAV-specific systems, such as CognitiveDrone and UAV-VLA \cite{lykov2025cognitivedrone,10.5555/3721488.3721725}, highlighted aerial navigation and satellite-informed planning. Additional contributions, including OneTwoVLA \cite{lin2025onetwovlaunifiedvisionlanguageactionmodel}, addressed adaptive reasoning–action switching and abstraction of heterogeneous control spaces.

In contrast to the aforementioned methods, efforts toward systematization and platformization have begun to consolidate these advances. Gemini Robotics \cite{geminiroboticsteam2025geminiroboticsbringingai} and Agentic Robot \cite{yang2025agenticrobotbraininspiredframework} positioned LLMs as central orchestrators of embodied pipelines, while $\pi$0.5 and fast \cite{intelligence2025pi05visionlanguageactionmodelopenworld,pertsch2025fastefficientactiontokenization} targeted open-world scalability and efficient tokenization. Supporting works, including VLA Model–Expert Collaboration and LLaRA \cite{xiang2025vlamodelexpertcollaborationbidirectional,li2024llara}, explored collaborative mechanisms and auxiliary tasks for improved VLM-to-VLA transfer. LLM-based reasoning in VLA has progressed from semantic intermediaries to interactive and hierarchical planning, cross-modal extensions, and integrated platforms.

Although LLM-based reasoning in VLA has evolved from semantic intermediaries to interactive and hierarchical planners, cross-modal extensions, and integrated platforms. Nonetheless, persistent challenges remain, including hallucination control, multimodal alignment, reasoning stability, and real-time safety. Table~\ref{tab:merged-vla}(B) summarizes representative studies and their contributions.

\subsubsection{Autoregressive Trajectory Generation and Visual Alignment Modeling}


Autoregressive trajectory modeling strengthens perception-action mapping while ensuring vision-language semantic alignment. These models decode motion trajectories or control tokens conditioned on multimodal observations, offering unified mechanisms for grounded instruction following and action execution~\cite{wei2024occllamaoccupancylanguageactiongenerativeworld,chen2025combatvlaefficientvisionlanguageactionmodel, li2025jarvisvlaposttraininglargescalevision}.

Early work such as LATTE~\cite{bucker2022lattelanguagetrajectorytransformer} showed the feasibility of mapping language directly to trajectories, inspiring multimodal extensions. With large-scale pretraining, VIMA~\cite{jiang2023vimageneralrobotmanipulation} and InstructRL~\cite{liu2023instructionfollowingagentsmultimodaltransformer} demonstrated that joint tokenization of language, vision, and action supports strong cross-task generalization, though often only in simulation. Meanwhile, MOO~\cite{stone2023openworldobjectmanipulationusing} and GPT-based approaches~\cite{Kwon_2024} leveraged pretrained vision–language backbones for open-world generalization and lightweight trajectory generation, suggesting semantic priors can reduce reliance on robot-specific pretraining.


A second line of work explores video prediction and world modeling. GR-1/2~\cite{wu2024unleashing,cheang2024gr2generativevideolanguageactionmodel} transferred video generation pretraining to robotics, while CronusVLA~\cite{li2025cronusvlatransferringlatentmotion} and WorldVLA~\cite{cen2025worldvla} improved temporal consistency. TraceVLA~\cite{zheng2025tracevlavisualtraceprompting} and Uni-NaVid~\cite{zhang2025uninavidvideobasedvisionlanguageactionmodel} further introduced long-horizon prompting, collectively shifting from short-horizon decoding to predictive environment modeling.


Autoregression-based methods have been adapted to diverse robot embodiments, from quadruped locomotion to bimanual manipulation, demonstrating the flexibility of Vision-Language-Action frameworks~\cite{10.1007/978-3-031-72652-1_21,gbagbe2024bivlavisionlanguageactionmodelbasedbimanual,zhao2023learningfinegrainedbimanualmanipulation}. Large-scale efforts such as OpenVLA~\cite{kim2024openvlaopensourcevisionlanguageactionmodel,kim2025finetuningvisionlanguageactionmodelsoptimizing} further highlight cross-platform generalization and efficient adaptation, while latent motion token approaches~\cite{ye2025latent,chen2025motolatentmotiontoken} point toward lightweight pretraining strategies.

Beyond manipulation, autoregressive trajectory generation has extended to autonomous driving, where recent models achieve closed-loop control by aligning vision and language with trajectory prediction, often without HD maps or LiDAR~\cite{bartoccioni2025vavimvavamautonomousdriving,renz2025simlingovisiononlyclosedloopautonomous,fu2025orionholisticendtoendautonomous}. Similar principles have been applied to mobile manipulation and UAV planning~\cite{Wu_2025_CVPR,serpiva2025racevlavlabasedracingdrone}, underscoring the versatility of these methods across robotic platforms.


Researchers have also extended autoregressive frameworks toward fine-grained sensing and richer modalities. Recent models emphasize precise manipulation with robust pretraining pipelines~\cite{goyal2024rvt2learningprecisemanipulation,deng2025graspvlagraspingfoundationmodel,lu2023unifiedio2scalingautoregressive,fu2024incontextimitationlearningnexttoken,niu2025pretrainingautoregressiveroboticmodels}, while tactile–language–action integration~\cite{hao2025tlatactilelanguageactionmodelcontactrich,zhang2025vtlavisiontactilelanguageactionmodelpreference} enables contact-rich interaction. Parallel efforts exploit 3D/4D perception to embed spatial structure into autoregressive decoding~\cite{singh2025ogvla3dawarevisionlanguage,zhang2025vla,li2025pointvlainjecting3dworld}, further broadening the multimodal landscape.


Autoregression-based trajectory generation has progressed from direct language-to-trajectory mapping toward a broad ecosystem that spans multimodal pretraining, video-driven world modeling, embodiment-specific architectures, and cross-modal sensing (see Table~\ref{tab:merged-vla} (C)). These advances showcase the scalability and versatility of autoregression as a unifying mechanism for VLA. Nonetheless, challenges remain in long-horizon stability, semantic grounding under noisy inputs, and efficient deployment on physical robots. Future work should prioritize robust closed-loop integration between predictive modeling and low-level control, and explore synergies between autoregressive policies and higher-level reasoning modules such as LLM planners, moving closer to reliable, general-purpose embodied intelligence.

\subsubsection{Structural Optimization and Efficient Inference Mechanisms in Autoregressive VLA}

In autoregressive VLA research, structural optimization and efficient inference are critical for enabling scalable deployment and real-time control. Beyond accuracy, the central challenge is how to reduce computational redundancy, shorten inference latency, and maintain robustness across diverse robotic settings.

One important direction is hierarchical and modular optimization. Early work such as HiP~\cite{ajay2023compositionalfoundationmodelshierarchical} demonstrated that decomposing tasks into symbolic planning, video prediction, and action execution enables long-horizon reasoning with autoregressive models. Follow-up designs—ranging from efficient observation backbones and action chunking to trajectory-aware attention and frequency separation~\cite{haldar2024bakuefficienttransformermultitask,ma2024actraoptimizedtransformerarchitecture,zhang2024hirt}—further showed that modular structures can significantly reduce computation while preserving generalization.


Another line of work emphasizes dynamic and adaptive inference. Frameworks like DeeR-VLA~\cite{NEURIPS2024_67b0e7c7} enable early termination of decoding based on task complexity, while token-efficient designs such as FAST~\cite{pertsch2025fastefficientactiontokenization} compress long action sequences into variable-length tokens. Together, these approaches show how adaptive computation can improve real-time responsiveness with minimal loss in accuracy.

A third category emphasizes lightweight compression and parallelization. Quantization and layer-skipping methods~\cite{wang2025bitvla1bitvisionlanguageactionmodels,zhang2025molevladynamiclayerskippingvision} cut precision and dynamically activate only subsets of layers, significantly reducing computation. In parallel, decoding and redundancy-reduction strategies~\cite{song2025acceleratingvisionlanguageactionmodelintegrated,duan2025fastecotefficientembodied} accelerate inference without retraining, highlighting how architectural compression complements adaptive inference.


Compression and parallelization methods span quantization and layer-skipping~\cite{wang2025bitvla1bitvisionlanguageactionmodels,zhang2025molevladynamiclayerskippingvision}, which substantially reduce computation, as well as parallel decoding and redundancy-reduction strategies~\cite{song2025acceleratingvisionlanguageactionmodelintegrated,duan2025fastecotefficientembodied}, which accelerate inference without retraining.

Efficiency has also been pursued through sensor fusion and temporal reuse. Domain-specific optimizations such as voxelized spatial modeling~\cite{qu2025spatialvlaexploringspatialrepresentations}, adaptive key–value caching~\cite{xu2025vlacacheefficientvisionlanguageactionmodel}, and perception adaptation~\cite{jones2025sightfinetuninggeneralistrobot} reduce redundant computation while improving robustness.

Notably, several works integrate efficiency with multimodal reasoning. OTTER~\cite{huang2025ottervisionlanguageactionmodeltextaware} injects language-awareness into vision encoding, while ChatVLA~\cite{zhou2025chatvlaunifiedmultimodalunderstanding} employs staged coupling with mixture-of-experts routing. Other advances—ranging from diffusion-based goal generation~\cite{li2025gr}, quantization~\cite{wang2025bitvla1bitvisionlanguageactionmodels}, to hierarchical feedback for ultra-long horizons~\cite{yang2025lohovlaunifiedvisionlanguageactionmodel}—illustrate how architectural refinements can balance efficiency and scalability.

In summary, structural optimization and efficient inference in autoregressive VLA models have evolved from early hierarchical decomposition strategies to adaptive computation, lightweight compression, caching, and multimodal-aware integration (see Table~\ref{tab:merged-vla}(D)). These methods address long-sequence dependency and computational redundancy, delivering notable gains in both benchmarks and real-world deployment. Looking forward, future research should pursue hardware-aware co-optimization, intelligent scheduling, and robust safety mechanisms to ensure scalable and reliable progress toward general-purpose embodied intelligence.

\subsubsection{Discussion}
\textbf{Innovations}
Autoregression-based models have driven major innovations in Vision-Language-Action research by unifying multimodal perception, language reasoning, and sequential action generation within scalable Transformer architectures. They support generalist agents capable of cross-task generalization, enable semantic planning through LLM integration, and extend trajectory generation to long-horizon and multimodal settings, structural optimizations such as token compression, parallel decoding, and quantization improve efficiency for real-world deployment. 

\textbf{Limitations}
Autoregressive decoding introduces error accumulation and latency, multimodal alignment can be brittle under noisy or incomplete inputs, and scaling large models requires prohibitive computational resources and data. Moreover, reasoning-driven approaches still face challenges of hallucination, stability, and interpretability, while efficiency mechanisms often trade off accuracy or generality. Addressing these issues will require tighter coupling between reasoning and control, robustness under real-world uncertainty, and hardware-aware optimization strategies to balance scalability with practical deployment.

\subsection{Diffusion-based Models in Vision-Language-Action Research}

Diffusion (including flow matching, VAE, etc.) models have emerged as a transformative paradigm in generative artificial intelligence, showing remarkable potential within Vision-Language-Action (VLA) frameworks for embodied intelligence. In this subsection, we review the evolution of diffusion models in VLA systems, focusing on three critical dimensions. 
Representative works are summarized in Table~\ref{tab:difussion}.

\begin{table*}[!htpb]
\centering
\caption{Representative works on Diffusion-based Models.}
\label{tab:difussion}
\footnotesize
\begin{tabular}{p{3.5cm}p{1cm}p{12.5cm}}
\toprule
\textbf{Method / Model} & \textbf{Year} & \textbf{Key Innovation in Diffusion-based Models} \\
\midrule

\multicolumn{3}{l}{\textbf{(A) Diffusion Generalist VLA Methodologies}} \\
\midrule
SE(3)-DiffusionFields~\cite{urain2023se3diffusionfieldslearningsmoothcost} & 2023 & Extended diffusion to SE(3) poses, learning smooth costs for grasp and motion planning. \\
UPDP~\cite{du2023learninguniversalpoliciestextguided} & 2023 & Framed decision-making as video generation with images as interfaces and language guidance. \\
StructDiffusion~\cite{liu2023structdiffusionlanguageguidedcreationphysicallyvalid} & 2023 & Combined diffusion and object-centric transformers for language-guided 3D structure creation. \\
Diffusion Policy~\cite{chi2024diffusionpolicyvisuomotorpolicy} & 2024 & Modeled actions as conditional diffusion, outperforming behavioral cloning. \\
3D Diffuser Actor~\cite{ke20243ddiffuseractorpolicy} & 2024 & Embedded 3D scenes with conditional diffusion for trajectory generation. \\
AVDC~\cite{ko2024learning} & 2024 & Learned visuomotor policies from video via optical flow and motion reconstruction. \\
RDT-1B~\cite{ICLR2025_49f80e4d} & 2025 & Large-scale diffusion model for bimanual manipulation with temporal conditioning. \\
TUDP~\cite{niu2025timeunifieddiffusionpolicyaction} & 2025 & Unified diffusion across time with velocity fields and action discrimination. \\
CDP~\cite{ma2025cdprobustautoregressivevisuomotor} & 2025 & Improved coherence via historical action conditioning and caching mechanisms. \\
DD VLA~\cite{liang2025discretediffusionvlabringing} & 2025 & Modeled discretized action chunks with discrete diffusion and cross-entropy training. \\
TRETL~\cite{li2025taskreconstructionextrapolationpi0} & 2025 & Recombined task behaviors by averaging hidden states across demonstrated trajectories. \\
\midrule
\multicolumn{3}{l}{\textbf{(B) Diffusion-based Multimodal Architectural Fusion}} \\
\midrule
SuSIE~\cite{black2023zeroshotroboticmanipulationpretrained} & 2023 & Pre-trained image-edit diffusion models for zero-shot robot manipulation via goal image generation. \\
M-DiT~\cite{reuss2024multimodaldiffusiontransformerlearning} & 2024 & Unified multimodal tokens enabling flexible language–image–position goal conditioning. \\
CogACT~\cite{li2024cogactfoundationalvisionlanguageactionmodel} & 2024 & Introduced cognition modules with semantic graphs linking perception and behavior. \\
PERIA~\cite{NEURIPS2024_1f6af963} & 2024 & Jointly fine-tuned MLLMs and image editing models for reasoning and sub-goal planning. \\
Chain-of-Affordance~\cite{li2024improvingvisionlanguageactionmodelschainofaffordance} & 2024 & Parsed tasks into sequential affordance sub-goals with explicit perception–action pairs. \\
$\pi$0~\cite{black2024pi0visionlanguageactionflowmodel} & 2024 & Encoded video and language as latent tokens in observe–understand–execute loops. \\
Track2Act~\cite{bharadhwaj2024track2actpredictingpointtracks} & 2024 & Leveraged web videos to predict point tracks and map them to robot actions. \\
Dita~\cite{hou2025dita} & 2025 & Scalable diffusion Transformer denoising continuous actions directly. \\
Diffusion-VLA~\cite{wen2025diffusionvlageneralizableinterpretablerobot} & 2025 & Integrated self-generated reasoning with diffusion policies via symbolic intermediates. \\
ForceVLA~\cite{yu2025forcevla} & 2025 & Incorporated 6-axis force sensing with force-aware MoE fusion for real-time control. \\
\midrule
\multicolumn{3}{l}{\textbf{(C) Application Optimization and Deployment in Diffusion-Based VLA}} \\
\midrule
NoMaD~\cite{sridhar2023nomadgoalmaskeddiffusion} & 2023 & Unifies goal-directed navigation and oal-agnostic exploration in diffusion policy. \\
TinyVLA~\cite{wen2025tinyvlafastdataefficientvisionlanguageaction} & 2025 & Uses LoRA fine-tuning with only 5\% trainable parameters,  effectively reducing training costs. \\
SmolVLA~\cite{shukor2025smolvlavisionlanguageactionmodelaffordable} & 2025 & Deploys compact VLAs on consumer hardware with async inference. \\
VQ-VLA~\cite{wang2025vqvlaimprovingvisionlanguageactionmodels} & 2025 & Employs vector-quantized tokenizers to reduce sim-to-real gaps. \\
DexVLG~\cite{he2025dexvlgdexterousvisionlanguagegraspmodel} & 2025 & Trains large-scale grasp model for dexterous hand grasping on DexGraspNet for zero-shot dexterity. \\
AC-DiT~\cite{chen2025ac} & 2025 & Adapts diffusion Transformer with multimodal mobility conditioning. \\
DexVLA~\cite{wen2025dexvlavisionlanguagemodelplugin} & 2025 & Applies morphology curriculum learning for cross-robot adaptation. \\
MinD~\cite{chi2025mindunifiedvisualimagination} & 2025 & Combines video imagination with high-frequency diffusion policies. \\
Hume~\cite{song2025humeintroducingsystem2thinking} & 2025 & Integrates dual-system value-guided reasoning and fast denoising. \\
TriVLA~\cite{liu2025trivlatriplesystembasedunifiedvisionlanguageaction} & 2025 & Operates triple-system VLA with 36Hz dynamics and reasoning layers. \\
BYOVLA~\cite{hancock2025runtime} & 2025 & Adds runtime interventions to enhance robustness without retraining. \\
DreamVLA~\cite{dreamvla25} & 2025 & Employs dynamic self-reflective loops with CoT, error tokens, and expert layers.\\
GEVRM~\cite{zhang2025gevrmgoalexpressivevideogeneration} & 2025 & Uses IMC principles with text-guided video generation for robustness. \\
DriveMoE~\cite{yang2025drivemoemixtureofexpertsvisionlanguageactionmodel} & 2025 & Deploys scene/action-specialized Mixture-of-experts architectures for autonomous driving. \\
DreamGen~\cite{jang2025dreamgenunlockinggeneralizationrobot} & 2025 & Generates neural trajectories enabling humanoids to learn novel tasks. \\
EnerVerse~\cite{huang2025enerverse} & 2025 & Predicts embodied futures with multi-view autoregressive video diffusion. \\
VidBot~\cite{chen2025vidbot} & 2025 & Reconstructs 3D affordances from monocular videos with depth priors. \\
OFT~\cite{kim2025finetuningvisionlanguageactionmodelsoptimizing} & 2025 & Optimizes fine-tuning via parallel decoding and chunked actions. \\
TrackVLA~\cite{wang2025trackvlaembodiedvisualtracking} & 2025 & Uses shared LLM heads and diffusion for embodied visual tracking. \\
FP3~\cite{yang2025fp33dfoundationpolicy} & 2025 & Builds large-scale 3D foundation diffusion policy for manipulation. \\
GR00T N1~\cite{nvidia2025gr00tn1openfoundation} & 2025 & Provides humanoid foundation model with multimodal Transformer control. \\
ObjectVLA~\cite{zhu2025objectvlaendtoendopenworldobject} & 2025 & Enables open-world object manipulation without human demos. \\
SwitchVLA~\cite{li2025switchvlaexecutionawaretaskswitching} & 2025 & Models execution-aware task switching from state–context signals. \\
A0~\cite{xu2025a0} & 2025 & An Affordance-aware model decomposes tasks into spatial affordances and low-level execution. \\
LangToMo~\cite{ranasinghe2025pixelmotionuniversalrepresentation} & 2025 & Forecasts pixel motion as intermediate reasoning for actions. \\
Evo-0~\cite{lin2025evo} & 2025 & Injects 3D geometry features via visual geometry foundation modules. \\

\bottomrule
\end{tabular}
\end{table*}

\subsubsection{Diffusion Generalist VLA Methodologies}


Diffusion models integration into VLA systems shifts robotic action generation from deterministic regression to probabilistic generative policies. By formulating action generation as conditional denoising, diffusion-based methods naturally model diverse action distributions, enabling multiple valid trajectories from identical observations~\cite{chi2024diffusionpolicyvisuomotorpolicy, niu2025timeunifieddiffusionpolicyaction}.


A key trajectory has been the incorporation of richer representational structures. Geometry-aware methods embed SE(3) constraints into diffusion, extending beyond Euclidean spaces to jointly optimize grasping and motion in 3D environments~\cite{urain2023se3diffusionfieldslearningsmoothcost,ke20243ddiffuseractorpolicy}, thereby ensuring physically consistent actions. In parallel, reinterpreting policy learning as video generation~\cite{du2023learninguniversalpoliciestextguided,ko2024learning} leverages the temporal richness of video for long-horizon planning and cross-modal grounding.


Scaling efforts like RDT-1B~\cite{ICLR2025_49f80e4d} demonstrate trajectory-level diffusion with temporal and environmental conditioning for zero-shot generalization in bimanual manipulation. Temporal coherence is addressed through unified velocity fields across timesteps~\cite{niu2025timeunifieddiffusionpolicyaction} or historical conditioning with efficient caching for real-time deployment~\cite{ma2025cdprobustautoregressivevisuomotor}.

These advances mark three transitions: deterministic to probabilistic generation, Euclidean to geometry-aware representations, and supervised to self-supervised paradigms. This reframing as generative modeling enables multi-task generalization, few-shot adaptation, and natural language interfaces. Table~\ref{tab:difussion}(A) summarizes architectural choices and training strategies. However, temporal coherence remains fragile under dynamic environmental shifts.

\subsubsection{Diffusion-Based Multimodal Architectural Fusion}


The integration of Transformers in VLA systems drives unified modeling of vision, language, and action within single frameworks, moving beyond modular pipelines to capture intricate interdependencies in embodied intelligence.

Within this shift, combining Transformers with diffusion models has proven especially transformative, as attention mechanisms naturally complement generative modeling. Large-scale frameworks such as Dita~\cite{hou2025dita} and Diffusion Transformer Policy~\cite{hou2025diffusiontransformerpolicy} show that scaling attention-based architectures beyond small action heads substantially improves continuous action modeling, with self-attention inductive biases aligning well with the compositionality of robotic behaviors.


The central challenge lies not in scaling architectures but in fusing heterogeneous modalities while preserving their distinct properties. Vision, language, and proprioception differ in temporal granularity, semantics, and processing needs—creating opportunities for richer context but also risks of diluting modality-specific strengths. To address this, token-space alignment strategies such as M-DiT~\cite{reuss2024multimodaldiffusiontransformerlearning} map diverse signals into unified representations, enabling conditional diffusion Transformers to flexibly support arbitrary combinations of goals and observations, a key step toward general-purpose robotics.


Domain-specific designs like ForceVLA~\cite{yu2025forcevla} treat force sensing as a first-class modality, using force-aware mixture-of-experts to integrate tactile feedback with visual-language embeddings, significantly improving contact-rich manipulation.

Recent advances integrate explicit reasoning within diffusion policies. Diffusion-VLA~\cite{wen2025diffusionvlageneralizableinterpretablerobot} introduces Self-Generated Reasoning modules producing symbolic representations, while CogACT~\cite{li2024cogactfoundationalvisionlanguageactionmodel} leverages semantic scene graphs, unifying perception, reasoning, and control.


Pre-trained model leveraging includes repurposing image editing models for zero-shot manipulation~\cite{black2023zeroshotroboticmanipulationpretrained} and joint fine-tuning strategies like PERIA~\cite{NEURIPS2024_1f6af963}. Structured decomposition through Chain-of-Affordance~\cite{li2024improvingvisionlanguageactionmodelschainofaffordance} and flow-graph approaches like $\pi$0~\cite{black2024pi0visionlanguageactionflowmodel} outperform end-to-end methods in complex environments.

Taken together, these developments (Table~\ref{tab:difussion}(B)) reveal a field in transition—from monolithic architectural adaptations toward cognitively inspired frameworks that integrate structured reasoning, multi-sensory inputs, and explicit knowledge representation. This shift signals a move beyond purely data-driven end-to-end learning toward more interpretable and generalizable designs, though progress remains constrained by high computational demands and limited dataset diversity.

\subsubsection{Application Optimization and Deployment in Diffusion-Based VLA}

The transition from laboratory prototypes to real-world deployment remains one of the most formidable challenges for diffusion-based Vision-Language-Action (VLA) systems. Meeting this challenge requires advances across three interconnected fronts: efficiency, adaptability, and robustness. Recent work demonstrates that rather than scaling model size indiscriminately, progress now hinges on principled optimization strategies, cognitive-inspired architectures, and practical deployment mechanisms.


Efficiency optimization has become a central theme. While diffusion models are resource-intensive, lightweight designs such as TinyVLA and SmolVLA~\cite{wen2025tinyvlafastdataefficientvisionlanguageaction,shukor2025smolvlavisionlanguageactionmodelaffordable} show that pre-trained backbones with parameter-efficient tuning (e.g., LoRA) can cut training costs to single-GPU scales without sacrificing performance. Complementary strategies like VQ-VLA~\cite{wang2025vqvlaimprovingvisionlanguageactionmodels}, which employs vector-quantized action tokenizers to narrow the sim-to-real gap, illustrate how efficiency gains can align with robustness. Collectively, these works reflect a paradigm shift toward “intelligent sparsity,” prioritizing performance-per-computation over brute-force scaling.


At the same time, task adaptability has become a defining feature of advanced VLA systems. In dexterous manipulation, large-scale curated datasets such as DexVLG~\cite{he2025dexvlgdexterousvisionlanguagegraspmodel} enable strong zero-shot performance, while in mobile manipulation, frameworks like AC-DiT~\cite{chen2025ac} unify perception and actuation through mobility-to-body conditioning. Overall, the trend is toward balancing general-purpose architectures with deep domain specialization, embedding task-specific inductive biases while retaining broad multimodal capabilities.

Architectural innovation represents the next frontier. Dual- and triple-system designs, such as MinD and TriVLA~\cite{chi2025mindunifiedvisualimagination,liu2025trivlatriplesystembasedunifiedvisionlanguageaction}, demonstrate how cognitive principles can be operationalized in robotics. MinD integrates low-frequency video prediction for strategic planning with high-frequency diffusion policies for reactive control, while TriVLA explicitly separates vision language reasoning, dynamics perception, and policy learning into coordinated modules. Operating at interactive frequencies (e.g., 36Hz), these cognitively inspired architectures not only improve task performance but also enhance system interpretability and maintainability—key requirements for industrial deployment.

Beyond efficiency and design, runtime robustness has emerged as a decisive factor for real-world adoption. Lightweight intervention strategies like BYOVLA~\cite{hancock2025runtime} dynamically edit irrelevant visual regions at inference time without fine-tuning, mitigating robustness failures in unpredictable environments. Meanwhile, self-reflective architectures such as DreamVLA~\cite{dreamvla25} introduce hierarchical error handling with reasoning-enhanced modules, error-aware layers, and expert adapters. Together, these strategies illustrate a shift toward “defensive AI,” emphasizing resilience and reliability as much as raw task performance.

The application landscape of diffusion-based VLA systems has rapidly expanded. In autonomous driving, DriveMoE~\cite{yang2025drivemoemixtureofexpertsvisionlanguageactionmodel} employs scene- and skill-specialized mixtures of experts to achieve state-of-the-art closed-loop control, while in humanoid robotics, DreamGen~\cite{jang2025dreamgenunlockinggeneralizationrobot} leverages video world models to generalize from single-task teleoperation to dozens of novel behaviors. EnerVerse~\cite{huang2025enerverse} and VidBot~\cite{chen2025vidbot} extend this paradigm by predicting embodied futures through autoregressive video diffusion and affordance learning, underscoring the potential of video-centric world models for planning. These advances highlight a transition from task-specific prototypes to versatile, domain-spanning systems.

Ambitious efforts toward foundation models further underscore the field’s trajectory. FP3~\cite{yang2025fp33dfoundationpolicy} introduces a large-scale 3D policy model pre-trained on 60,000 trajectories, while GR00T N1~\cite{nvidia2025gr00tn1openfoundation} integrates multimodal Transformer architectures into a humanoid foundation system. Like large language models in NLP, these approaches aim to provide general-purpose priors for robotics, though they must also address safety, real-time control, and physical reliability—challenges less pronounced in text-based domains.

Generalization and fine-tuning strategies remain critical for advancing diffusion-based VLA systems toward real-world deployment. Recent research highlights multiple complementary directions: ObjectVLA and SwitchVLA~\cite{zhu2025objectvlaendtoendopenworldobject,li2025switchvlaexecutionawaretaskswitching} demonstrate the feasibility of open-world object manipulation and execution-aware task switching, emphasizing flexibility in dynamic environments. In parallel, approaches such as LangToMo and Evo-0~\cite{ranasinghe2025pixelmotionuniversalrepresentation,lin2025evo} introduce novel intermediate representations and geometry-aware plug-in modules, showing that structured perceptual priors can significantly enhance adaptability across tasks. On the optimization front, systematic fine-tuning frameworks like OFT~\cite{kim2025finetuningvisionlanguageactionmodelsoptimizing} integrate techniques including parallel decoding, action chunking, and continuous representation learning, moving the field from exploratory proofs-of-concept toward an engineering discipline. 

Collectively, these strategies illustrate that achieving robust generalization requires architectural innovation, efficient model design, adaptive task specialization, cognitively inspired architectures, and robust runtime strategies, as summarized in Table~\ref{tab:difussion} (C). Nonetheless, challenges persist: safety-critical scenarios is still underdeveloped. Bridging these gaps is essential to transition from experimental prototypes to reliable, general-purpose robotic systems.

The application of diffusion models in VLA systems is evolving toward greater efficiency, robustness, and universality. From foundational action generation modeling to complex multimodal fusion and practical deployment optimization, a comprehensive technological framework has emerged. There are still issues need to be solved, future development trends will continue to address key challenges including model efficiency enhancement, generalization capability improvement, and practical deployment performance optimization.

\subsubsection{Discussion}
\textbf{Innovations}
Diffusion-based models fundamentally reframing robotic control as a generative modeling problem. They enable probabilistic action generation, multimodal architectural fusion, and cognitively inspired deployment strategies, advancing beyond deterministic and modular pipelines. These approaches improve trajectory diversity, geometric grounding, and reasoning integration. Besides, efficiency-focused designs such as TinyVLA and SmolVLA make real-world deployment increasingly feasible. 

\textbf{Limitations}
However, due to the fact that temporal coherence in dynamic environments is still fragile, large-scale diffusion models demand prohibitive computational resources and datasets, and safety-critical reliability under adversarial or uncertain conditions is underexplored. Furthermore, while multimodal fusion enriches representation, it risks diluting modality-specific strengths, and domain-specialized adaptations may reduce transferability. Addressing these challenges will require more efficient and robust training paradigms, richer safety-aware evaluation standards, and closer alignment between foundation-scale modeling and practical deployment constraints.

\subsection{Reinforcement-based Fine-Tune Models in Vision-Language-Action Research}

\subsubsection{Reinforcement-based Fine-Tune Strategies in VLA Research}

\begin{table*}[!htpb]
\centering
\caption{Evolution of Reinforcement-based Models in VLA Research and Key Innovations.}
\label{table:reinforcement}
\footnotesize
\begin{tabular}{p{3.5cm}p{1cm}p{12.5cm}}
\toprule
\textbf{Method} & \textbf{Year} & \textbf{VLA Innovation with Reinforcement Learning} \\
\midrule

VIP~\cite{ma2023vipuniversalvisualreward} & 2023 & Generates dense reward functions from visual pretraining for unseen tasks. \\
LIV~\cite{ma2023livlanguageimagerepresentationsrewards} & 2023 & Learns joint vision–language rewards from action-free videos with text. \\
PR2L~\cite{chen2024visionlanguagemodelsprovidepromptable} & 2024 & Leverages VLM world knowledge with RL for robot manipulation. \\
ALGAE~\cite{peng2024adaptivelanguageguidedabstractioncontrastive} & 2024 & Introduces language-guided abstractions to explain RL-driven behavior. \\
GRAPE~\cite{zhang2024grape} & 2024 & Aligns VLAs at trajectory level, modeling rewards from successes and failures. \\
RLDG~\cite{xu2024rldgroboticgeneralistpolicy} & 2024 & Uses RL to generate training data for generalist policy finetuning. \\
ELEMENTAL~\cite{chen2025elementalinteractivelearningdemonstrations} & 2025 & Learns reward proxies from demonstrations via VLM-based semantic mapping. \\
NaVILA~\cite{cheng2025navilaleggedrobotvisionlanguageaction} & 2025 & Integrates multimodal VLA with RL for quadruped navigation on complex terrain. \\
SafeVLA~\cite{zhang2025safevlasafetyalignmentvisionlanguageaction} & 2025 & Constrains RL alignment to prevent risky actions while maintaining performance. \\
iRe-VLA~\cite{guo2025improvingvisionlanguageactionmodelonline} & 2025 & Combines SFT stability with RL exploration to boost VLA capability. \\
ReinboT~\cite{zhang2025reinbotamplifyingrobotvisuallanguage} & 2025 & Enhances Robot GPT with RL to maximize rewards and data quality use. \\
ConRFT~\cite{chen2025conrft} & 2025 & Blends offline BC, Q-learning, and online consistency for safe efficient RL. \\
MoRE~\cite{zhao2025moreunlockingscalabilityreinforcement} & 2025 & Scales quadruped VLAs with RL and mixed-quality dataset finetuning. \\
SimpleVLA-RL~\cite{li2025simplevlarl} & 2025 & Trains from one trajectory using binary (0/1) rewards in online RL. \\
LeVERB~\cite{xue2025leverbhumanoidwholebodycontrol} & 2025 & Couples VLA reasoning with RL dynamics for humanoid whole-body control. \\
AutoVLA~\cite{zhou2025autovla} & 2025 & Employs CoT reasoning with RL optimization for autonomous driving. \\
RPD~\cite{julg2025refined} & 2025 & Distills student from VLA teacher using RL-based refinement. \\
RLRC~\cite{chen2025rlrcreinforcementlearningbasedrecovery} & 2025 & Compresses VLA via pruning, SFT+RL recovery, and quantization. \\
VLA-RL~\cite{lu2025vlarlmasterfulgeneralrobotic} & 2025 & Improves robustness with online RL and VLM-based reward modeling. \\
AutoDrive-R$^2$~\cite{yuan2025autodriver2incentivizingreasoningselfreflection} & 2025 & Enhances autonomous driving with CoT reasoning and RL self-reflection. \\
ReWiND~\cite{zhang2025rewindlanguageguidedrewardsteach} & 2025 & Uses language-conditioned rewards for offline RL from small demos. \\
Embodied-R1~\cite{yuan2025embodiedr1reinforcedembodiedreasoning} & 2025 & Trains with reinforced fine-tuning and multi-task reward curriculum. \\
IRL-VLA~\cite{jiang2025irlvlatrainingvisionlanguageactionpolicy} & 2025 & Applies inverse RL in a reward world model within a VLA framework. \\
ThinkAct~\cite{huang2025thinkactvisionlanguageactionreasoningreinforced} & 2025 & Bridges reasoning and execution with reinforced visual latent planning. \\

\bottomrule
\end{tabular}
\end{table*}

Reinforcement-based Vision-Language-Action (VLA) methods integrate vision–language foundation models with reinforcement learning to enhance perception, reasoning, and decision-making. By leveraging visual and linguistic inputs, these methods generate context-aware actions within interactive and dynamic environments. They have emerged as a critical research direction for advancing autonomous driving, robotics, and broader embodied AI systems. 
Recent advances demonstrate that reinforcement-based VLA approaches can incorporate human feedback, adapt to novel tasks, and outperform purely supervised paradigms. The progression of these efforts is summarized in Table~\ref{table:reinforcement}.

Earlier methods improved robot manipulation skills using large-scale human video datasets or robot manipulation datasets by introducing reinforcement reward strategies \cite{ma2023vipuniversalvisualreward, ma2023livlanguageimagerepresentationsrewards, chen2024visionlanguagemodelsprovidepromptable, guo2025improvingvisionlanguageactionmodelonline}. These methods aimed to investigate the promptability of pre-trained Vision-Language Models (VLMs) in reinforcement learning, showing that even frozen models can support efficient downstream policy training through prompt embedding learning. 
VIP \cite{ma2023vipuniversalvisualreward} derives a self-supervised goal-conditioned value function independent of actions, generating smooth embeddings that implicitly evaluate value via embedding distance. 

Similar to other reinforcement fine-tuning approaches, some methods use language and images to jointly generate reward proxies and obtain cross-modal state language representations through self-supervised contrastive training. These methods emphasize the transferability of reward-aware representations, enabling applications in robot learning under sparse rewards or complex language instructions \cite{ma2023livlanguageimagerepresentationsrewards, guo2025improvingvisionlanguageactionmodelonline, lu2025vlarlmasterfulgeneralrobotic}.

Furthermore, some approaches primarily optimize reward functions or loss functions to improve policy learning \cite{chen2025elementalinteractivelearningdemonstrations, peng2024adaptivelanguageguidedabstractioncontrastive}. These methods use language models as intermediaries for reward function design, learning reward proxies through human demonstrations and VLM semantic mapping. This approach simplifies reward engineering, while generalization and interpretability can be further optimized with reinforcement learning from human feedback (RLHF). For example, Elemental demonstrates the ability to rapidly customize task requirements and efficiently learn from limited samples in complex manipulation tasks.
SafeVLA \cite{zhang2025safevlasafetyalignmentvisionlanguageaction} explores VLA from a safety perspective, addressing the risks of deploying VLAs in open environments. It proposes a constrained-learning alignment mechanism to prevent high-risk behaviors while maintaining task performance. The method incorporates a safety critic network into the VLA architecture to estimate risk levels and employs the Constrained Policy Optimization (CPO) framework to maximize policy rewards while ensuring that security loss remains below a predefined threshold. SafeVLA significantly reduces risk events in multitask testing—including manipulation, navigation, and handling—particularly in scenarios where ambiguous natural language instructions increase policy uncertainty, thereby demonstrating superior safety and stability. This work provides an essential safety mechanism for deploying VLA models in real-world applications.

Unlike the aforementioned robot arm VLA models, researchers have also investigated VLA frameworks for quadruped and humanoid robots. Using natural language navigation instructions, these robots emphasize trajectory prediction, target description, obstacle avoidance, and related tasks. For example, NaVILA \cite{cheng2025navilaleggedrobotvisionlanguageaction} fine-tunes a VLA model with a single-stage reinforcement learning (RL) policy to output continuous control commands, enabling adaptation to complex terrain and dynamically changing language instructions. In contrast, MoRE \cite{zhao2025moreunlockingscalabilityreinforcement} integrates multiple low-rank adaptive modules as distinct experts into a dense multimodal large language model (MLLM), forming a sparsely activated hybrid expert model that is subsequently trained as a Q-function using reinforcement learning objectives. LeVERB \cite{xue2025leverbhumanoidwholebodycontrol} extends this line of research by proposing a hierarchical VLA framework for whole-body control (WBC) of humanoid robots. Similar to NaVILA, LeVERB couples vision–language processing with dynamics-level action processing, where reinforcement learning strategies translate potential vocabularies into high-frequency dynamic control commands, enabling complex whole-body task execution.

Offline reinforcement learning has proven effective for deriving robust policy models from mixed-quality datasets. ReinboT \cite{zhang2025reinbotamplifyingrobotvisuallanguage} exemplifies this approach by applying the principle of maximizing cumulative rewards via RL. It enhances understanding of data quality distribution by predicting dense rewards that capture subtle differences in operational tasks, thereby enabling robots to generate more robust decision actions guided by long-term benefits. Online reinforcement learning methods have also been extensively explored in the VLA domain. For instance, SimpleVLA-RL \cite{li2025simplevlarl} employs only a single trajectory and a binary outcome-level reward (0/1) to train a VLA model. This method avoids reliance on dense supervision or large-scale behavior cloning datasets, but achieves performance comparable to full trajectory supervised fine-tuning (SFT) with simulating rule-based reward signals in the environment. Recognizing the limitations of using only offline or online strategies, ConRFT \cite{chen2025conrft} introduces a hybrid strategy that combines both. Its offline policy incorporates behavior cloning with Q-learning to extract policies from limited demonstrations and stabilize value estimation, while its online policy introduces consistency goals and artificial intervention mechanisms to steadily improve policy performance, ensuring safe exploration and sample efficiency throughout training.

In the autonomous driving domain, VLA models also leverage reinforcement learning to enhance driving performance in previously unseen scenarios \cite{yuan2025autodriver2incentivizingreasoningselfreflection}. AutoVLA \cite{zhou2025autovla} exemplifies this direction by introducing an autoregressive generation model equipped with reasoning and action capabilities. It first processes visual inputs and language instructions, then applies reasoning fine-tuning to produce discrete, feasible actions that can be reconstructed into continuous trajectories. This model employs two fine-tuning steps—Chain-of-Thought Reasoning and Group Relative Policy Optimization—achieving state-of-the-art performance.

Notably, different from existing models that require enormous numbers of parameters, resulting in high computational and memory demands, some researchers have investigated efficiency strategies such as quantization, pruning, and knowledge distillation within reinforcement learning–based VLAs, often combined with algorithms such as Proximal Policy Optimization (PPO) \cite{schulman2017proximalpolicyoptimizationalgorithms}. For instance, RPD \cite{julg2025refined} distills a student model from a VLA teacher model to increase inference speed, while RLRC \cite{chen2025rlrcreinforcementlearningbasedrecovery} introduces a novel compression framework composed of structured pruning, SFT- and RL-based performance recovery, and quantization. These approaches reduce memory usage and improve inference throughput while preserving the task success rate of the original VLA.

\subsubsection{Discussion}
\textbf{Innovations}
Reinforcement-based VLA fine-tune strategies using visual and language signals to generate dense, transferable reward proxies, and combining offline behavior cloning with online reinforcement learning stabilizes policy optimization and enhances generalization. Safety-focused approaches also represent an important advancement by integrating constrained optimization to reduce high-risk actions in open-world deployment. Furthermore, extensions to quadruped, humanoid, and autonomous driving tasks highlight the versatility of reinforcement-driven VLA across diverse robotic embodiments.

\textbf{Limitations}
Despite these advances, reward of reinforcement-based VLA engineering often remains indirect or noisy, leading to suboptimal learning; training stability can be hindered by the interplay between supervised fine-tuning and exploration; and scaling to high-dimensional, real-world environments is computationally expensive, requiring substantial hardware and data resources. Additionally, while safety-aware strategies have been proposed, ensuring reliable generalization under ambiguous or adversarial instructions remains an open challenge. Addressing these issues will require more efficient reward representation, robust sample-efficient training paradigms, and richer evaluation benchmarks that capture both safety and reasoning capabilities.

\subsection{Other Advanced Researches}

\begin{table*}[!htpb]
\centering
\caption{Evolution of Hybrid Architectures and Specialized Approacher in Vision-Language-Action Research.}
\label{table:hybrid_architectures}
\footnotesize
\begin{tabular}{p{3.5cm}p{1cm}p{12.5cm}}
\toprule
\textbf{Method} & \textbf{Year} & \textbf{Key Innovation in Hybrid and Specialized Architecture} \\
\midrule
\multicolumn{3}{l}{\textbf{(A) Hybrid Architectures and Multi-Paradigm Integration in VLA}} \\
\midrule

VRB~\cite{bahl2023affordanceshumanvideosversatile} & 2023 & Learns affordances from human videos to enable complex robot interactions. \\
YAY Robot~\cite{shi2024yellrobotimprovingonthefly} & 2024 & Incorporates real-time natural language corrections into robot training. \\
HybridVLA~\cite{liu2025hybridvlacollaborativediffusionautoregression} & 2025 & Combines diffusion-based trajectories with autoregressive reasoning in one framework. \\
RationalVLA~\cite{song2025rationalvlarationalvisionlanguageactionmodel} & 2025 & Links high-level reasoning with low-level policies via latent embeddings. \\

OpenHelix~\cite{cui2025openhelix} & 2025 & Provides standardized hybrid VLA designs through large-scale empirical studies. \\
EgoVLA~\cite{yang2025egovlalearningvisionlanguageactionmodels} & 2025 & Transfers human wrist/hand actions from video into robot actions. \\
ACTLLM~\cite{bi2025actllmactionconsistencytuned} & 2025 & Enforces action consistency to align perception with executable actions. \\
TUDP~\cite{niu2025timeunifieddiffusionpolicyaction} & 2025 & Builds a time-unified diffusion policy with velocity and discrimination signals. \\

\midrule
\multicolumn{3}{l}{\textbf{(B) Advances in Multi-Modal Fusion and Spatial Understanding in VLA}} \\
\midrule

CLIPort~\cite{shridhar2021cliportpathwaysroboticmanipulation} & 2021 & Separates “what” and “where” pathways with CLIP to generate action heatmaps. \\
3D-VLA~\cite{zhen20243dvla3dvisionlanguageactiongenerative} & 2024 & Integrates perception, language, and action via generative 3D world modeling. \\
ReKep~\cite{huang2024rekepspatiotemporalreasoningrelational} & 2024 & Leverages relational keypoint graphs for precise spatio-temporal reasoning. \\
RoboPoint~\cite{yuan2024robopointvisionlanguagemodelspatial} & 2024 & Predicts affordance maps as priors for downstream robotic planning. \\
GMLLMA~\cite{szot2024groundingmultimodallargelanguage} & 2024 & Adapts MLLMs across embodiments and action spaces for grounded reasoning. \\
BridgeVLA~\cite{li2025bridgevlainputoutputalignmentefficient} & 2025 & Aligns 3D observations to 2D heatmaps for sample-efficient action prediction. \\
GeoManip~\cite{tang2025geomanipgeometricconstraintsgeneral} & 2025 & Embeds geometric constraints to generalize actions without retraining. \\
TTF-VLA~\cite{liu2025ttfvlatemporaltokenfusion} & 2025 & Fuses past and current visual tokens for training-free inference improvement. \\
MemoryVLA~\cite{shi2025memoryvlaperceptualcognitivememoryvisionlanguageaction} & 2025 & Builds working memory with perceptual and cognitive tokens plus a memory bank. \\

\midrule
\multicolumn{3}{l}{\textbf{(C) Specialized Domain Adaptations and Applications in VLA}} \\
\midrule

GenAug~\cite{chen2023genaugretargetingbehaviorsunseen} & 2023 & Uses semantic data augmentation to retarget behaviors with minimal real data. \\
ROSIE~\cite{yu2023scalingrobotlearningsemantically} & 2023 & Applies diffusion-based augmentation to enrich manipulation datasets. \\
CoVLA~\cite{arai2024covlacomprehensivevisionlanguageactiondataset} & 2024 & Builds a large VLA dataset for AD with paired instructions and trajectories. \\
LeVERB~\cite{xue2025leverbhumanoidwholebodycontrol} & 2025 & Designs hierarchical policies for humanoid body control with robust sim-to-real transfer. \\
Helix~\cite{figure2024helix} & 2024 & Unifies humanoid control for manipulation, locomotion, and collaboration. \\
AutoRT~\cite{ahn2024autortembodiedfoundationmodels} & 2024 & Orchestrates robot fleets via observe–reason–execute with foundation models. \\
MoManipVLA~\cite{wu2025momanipvlatransferringvisionlanguageactionmodels} & 2025 & Adapts fixed-base VLAs to mobile robots through waypoint and motion optimization. \\

CubeRobot~\cite{wang2025cuberobot} & 2025 & Solves Rubik’s Cube with dual-loop VisionCoT and memory stream for high success. \\
EAV-VLA~\cite{wang2025exploringadversarialvulnerabilitiesvisionlanguageaction} & 2025 & Develops adversarial patch attacks to destabilize or redirect robot actions. \\

\midrule
\multicolumn{3}{l}{\textbf{(D) Foundation Models and Large-Scale Training in VLA}} \\
\midrule

R3M~\cite{nair2022r3muniversalvisualrepresentation} & 2022 & Learns compact visual representations via time-contrastive, video-language alignment. \\
CACTI~\cite{mandi2023cactiframeworkscalablemultitask} & 2023 & Establishes a scalable four-stage pipeline for multi-task robot learning in simulation and real-world. \\
VC-1~\cite{majumdar2024searchartificialvisualcortex} & 2024 & Studies pre-training data scale using 4,000+ hours of video with MAE-trained transformers. \\
DROID~\cite{khazatsky2025droidlargescaleinthewildrobot} & 2025 & Provides a large multimodal dataset of 150k+ trajectories across 1k+ objects and tasks. \\
ViSA-Flow~\cite{chen2025visa} & 2025 & Pre-trains generative models on action flows from large-scale human–object interaction videos. \\
Fine-tuned GMPs~\cite{zhang2024effectivetuningstrategiesgeneralist} & 2024 & Benchmarks fine-tuning strategies for generalist policies across action spaces and signals. \\
Robot CoT~\cite{chen2025training} & 2025 & Introduces lightweight chain-of-thought reasoning for embodied policies with 3× faster inference. \\
CAST~\cite{glossop2025castcounterfactuallabelsimprove} & 2025 & Enhances dataset diversity via counterfactual language and action generation. \\
RoboBrain~\cite{ji2025robobrainunifiedbrainmodel} & 2025 & Proposes a unified embodied foundation model for perception, reasoning, and planning. \\

\midrule
\multicolumn{3}{l}{\textbf{(E) Deployment of VLA Models: Efficiency, Safety, and Human–Robot Collaboration}} \\
\midrule

EdgeVLA~\cite{kscale2024evla} & 2024 & Achieves 6× faster inference by removing dependencies and using compact LLMs. \\
DeeR-VLA~\cite{yue2024deervladynamicinferencemultimodal} & 2024 & Cuts control cost with confidence-based early exit. \\
Dual Process VLA~\cite{han2024dualprocessvlaefficient} & 2024 & Splits reasoning and motor control across dual models for efficiency. \\
CEED-VLA~\cite{song2025ceedvlaconsistencyvisionlanguageactionmodel} & 2025 & Speeds up inference 4× with consistency distillation and early exit. \\
RoboMamba~\cite{liu2024robomambaefficientvisionlanguageactionmodel} & 2024 & Uses lightweight fusion for resource-limited deployment. \\
ReVLA~\cite{dey2025revlarevertingvisualdomain} & 2025 & Adapts across visual domains to improve robustness. \\
SAFE~\cite{gu2025safe} & 2025 & Proactively detects failures from internal VLA signals. \\
DyWA~\cite{lyu2025dywadynamicsadaptiveworldaction} & 2025 & Models dynamics for adaptive control in uncertain environments. \\
CrayonRobo~\cite{li2025crayonroboobjectcentricpromptdrivenvisionlanguageaction} & 2025 & Provides interpretable grounding via object-centric prompts. \\
cVLA~\cite{argus2025cvla} & 2025 & Enhances sim-to-real transfer with 2D waypoint prediction. \\
RTC~\cite{black2025realtimeexecutionactionchunking} & 2025 & Supports smooth asynchronous execution of chunked policies. \\

\bottomrule
\end{tabular}
\end{table*}

While autoregressive, diffusion, and reinforcement learning remain the foundational paradigms of VLA model design, the growing complexity and diversity of embodied tasks have spurred the development of approaches that transcend these boundaries. Current research progress can be organized into five key directions: hybrid architectures that integrate multiple generation paradigms, advanced multi-modal fusion for enhanced cross-modal and spatial understanding, specialized domain adaptations addressing task-specific challenges, foundation models and large-scale training paradigms that unify perception–reasoning–control at scale, and practical deployment strategies emphasizing efficiency, safety, and human–robot collaboration. The representative works are
summarized in Table~\ref{table:hybrid_architectures}.

\subsubsection{Hybrid Architectures and Multi-Paradigm Integration}
As embodied manipulation tasks continue to grow in diversity and complexity, relying on a single generation paradigm (whether autoregressive, diffusion, or reinforcement learning) often proves inadequate. Hybrid architectures have thus emerged as a promising solution, strategically combining multiple paradigms to exploit their complementary strengths. The central objective of this approach is to integrate the smoothness and physical consistency of continuous action generation, the precision of discrete reasoning, and the adaptability needed for dynamic, real-world environments. In doing so, hybrid systems lay the groundwork for more capable and versatile VLA models. 

A representative example is HybridVLA~\cite{liu2025hybridvlacollaborativediffusionautoregression}, which unifies diffusion-based continuous trajectory generation with autoregressive token-level reasoning in a single 7B-parameter framework. This design leverages diffusion processes to produce smooth and physically coherent motion, while retaining the contextual inference capacity inherent in autoregressive models.
The dual-system philosophy, inspired by cognitive science, has also been embraced in recent works. Fast-in-Slow\cite{chen2025fastinslowdualsystemfoundationmodel} operationalizes Kahneman’s dual-process theory by embedding a low-latency execution module within a slower yet cognitively richer VLM backbone. This enables real-time responsiveness while preserving high-level reasoning. Similarly, RationalVLA\cite{song2025rationalvlarationalvisionlanguageactionmodel} integrates vision–language reasoning with low-level manipulation policies through learnable latent embeddings, allowing the model to filter out infeasible commands and plan executable actions. 

Scaling hybrid designs has also shown considerable promise. Transformer-based Diffusion Policy~\cite{zhu2024scalingdiffusionpolicytransformer} demonstrates that billion-parameter architectures can effectively combine diffusion processes with attention mechanisms, surpassing conventional U-Net designs by capturing richer contextual dependencies for trajectory modeling. This trend points toward the next generation of VLA systems that embed autoregressive Transformers within diffusion-based planners, achieving greater context-awareness alongside higher-quality motion generation.

Beyond individual innovations, initiatives such as OpenHelix~\cite{cui2025openhelix} are moving toward the systematization of hybrid VLA design. Through large-scale empirical evaluation, OpenHelix benchmarks alternative reasoning–execution integration strategies and provides open-source implementations alongside design guidelines. This shift signals a maturation of the field, fostering reproducibility and standardization in hybrid VLA development. The progression of these efforts is summarized in Table~\ref{table:hybrid_architectures} (A), which outlines the key innovations driving hybrid VLA architectures.

\subsubsection{Advanced Multi-Modal Fusion and Spatial Understanding}

Achieving robust manipulation in complex environments requires more than straightforward cross-modal alignment; it calls for structured, task-aware fusion mechanisms capable of capturing fine-grained semantics and spatial relationships. Recent progress reflects a decisive shift away from early feature concatenation toward architectures that explicitly model geometry, affordances, and spatial constraints. These advances are propelling VLA models toward richer spatial grounding and more reliable action generation in unstructured, 3D-aware settings.

Early efforts such as CLIPort~\cite{shridhar2021cliportpathwaysroboticmanipulation} laid the foundation by disentangling visual processing into a “what” pathway for object identification and a “where” pathway for action localization. Leveraging CLIP-based representations, CLIPort generates pick-and-place heatmaps from paired image–language inputs, demonstrating the advantages of structured visual reasoning in language-conditioned manipulation. Building upon this foundation, subsequent research has emphasized 3D spatial understanding as a core competency. VoxPoser\cite{huang2023voxposercomposable3dvalue} introduces composable 3D value maps guided by large language models, dividing instruction interpretation into target understanding and action planning over voxelized scene representations. This modular design enhances generalization by cleanly separating semantic parsing from spatial reasoning. Similarly, 3D-VLA\cite{zhen20243dvla3dvisionlanguageactiongenerative} integrates autoregressive language modeling with diffusion-based action prediction in a generative 3D world model, achieving coherent unification of perception, language, and action modalities.

The challenge of multi-view perception has been tackled through unified representation learning. RoboUniView\cite{liu2024robouniviewvisuallanguagemodelunified} employs multi-view Transformer modules to fuse temporal and spatial cues, substantially improving 3D scene geometry understanding compared to single-view baselines. In contrast, BridgeVLA\cite{li2025bridgevlainputoutputalignmentefficient} projects 3D observations into multiple 2D views and predicts actions within a unified 2D heatmap space, to highlight the efficiency of compact yet spatially grounded representations.
To handle more demanding scenarios, specialized spatial reasoning methods have emerged. ReKep\cite{huang2024rekepspatiotemporalreasoningrelational} models spatio-temporal dependencies via relational keypoint graphs, excelling in precision-critical tasks. RoboPoint\cite{yuan2024robopointvisionlanguagemodelspatial} predicts affordance maps that highlight feasible interaction regions, providing essential perception priors for downstream planning. GeoManip~\cite{tang2025geomanipgeometricconstraintsgeneral} integrates symbolic geometric constraints to guide action generation without requiring task-specific retraining, thereby achieving strong out-of-distribution generalization.

Taken together, these works trace a clear trajectory: from early pathway-based 2D fusion to modular, 3D-aware architectures that unify spatial grounding, semantic reasoning, and action generation. As VLA systems increasingly operate in unconstrained, real-world environments, the capacity to reason explicitly about geometry and affordances will remain a decisive factor for achieving robust and generalizable manipulation. Table~\ref{table:hybrid_architectures} (B) summarizes this progression.

\subsubsection{Specialized Domain Adaptations and Applications}

The versatility of the VLA framework has enabled its expansion into specialized embodied domains that pose unique perceptual, reasoning, and control challenges. Such adaptations not only validate the generality of VLA principles but also reveal the architectural and algorithmic modifications necessary for domain-specific success. From safety-critical robotics to fully digital interaction, these innovations demonstrate the adaptability of VLA pipelines to varied operational contexts.

In safety-critical contexts such as autonomous driving, CoVLA~\cite{arai2024covlacomprehensivevisionlanguageactiondataset} presents the first large-scale VLA dataset tailored for this domain, comprising approximately 50,000 paired language instructions and driving trajectory videos across diverse urban scenarios. This work illustrates how vision–language reasoning can be coupled with continuous control policies for navigation and hazard avoidance.

The VLA paradigm has also been extended to graphical user interface (GUI) interaction, where perception-action loops operate in fully digital spaces. ShowUI~\cite{lin2024showuivisionlanguageactionmodelgui} adopts a Vision-Language-Action pipeline for processing on-screen elements and generating control sequences for actions such as clicking, dragging, and form filling. Its strong performance on GUI-Bench underscores the applicability of VLA principles to non-physical manipulation tasks.

Humanoid whole-body control has emerged as another challenging domain. LeVERB~\cite{xue2025leverbhumanoidwholebodycontrol} proposes a hierarchical architecture in which a vision–language policy learns latent action vocabularies from kinematic demonstrations, while a reinforcement-learned control layer produces low-level dynamics commands. This two-level design bridges the semantic–control gap, enabling robust sim-to-real transfer across more than 150 tasks. Similarly, Helix~\cite{figure2024helix} demonstrates that a single unified policy network can acquire diverse humanoid behaviors, from object manipulation to cross-robot collaboration, without task-specific retraining.

Specialized adaptations also target large-scale robot orchestration and mobile manipulation. AutoRT~\cite{ahn2024autortembodiedfoundationmodels} coordinates heterogeneous robot fleets via an observe–reason–execute framework that delegates strategic planning to VLMs such as PaLM-E and RT-2, while MoManipVLA~\cite{wu2025momanipvlatransferringvisionlanguageactionmodels} transfers fixed-base VLA models to mobile manipulation settings through waypoint-based trajectory generation and dual-layer motion optimization.

Other domain-specific innovations incorporate physical reasoning or task-specialized cognitive structures. Physically grounded VLA~\cite{gao2024physicallygroundedvisionlanguagemodels} embed modules for estimating stability and contact points, improving manipulation under complex physical constraints. CubeRobot~\cite{wang2025cuberobot} applies a dual-loop VisionCoT and Memory Stream design to Rubik’s Cube solving, achieving near-perfect success rates in low- and medium-complexity tasks, and strong performance in high-difficulty scenarios.

Overall, these domain-driven adaptations demonstrate the versatility of VLA architectures, as well as the importance of tailoring perception–reasoning–control pipelines to meet the specific demands of different operational contexts. They also reinforce the potential of VLA models as a unifying embodied intelligence framework spanning physical, digital, and hybrid environments. These specialized adaptations are summarized in Table~\ref{table:hybrid_architectures} (C), which highlights the architectural and algorithmic innovations enabling VLA systems to succeed across diverse embodied domains.

\subsubsection{Foundation Models and Large-Scale Training}

The rise of foundation models and large-scale training has reshaped the trajectory of VLA research, enabling unified perception–reasoning–control frameworks that generalize across tasks, embodiments, and environments. By leveraging massive multi-modal datasets and scalable architectures, this direction seeks to build generalist embodied agents with broad capabilities and efficient adaptation. Large-scale pre-training is increasingly becoming the backbone for next-generation VLA systems. 
Recent foundation models provide systematic researches in robotics, covering vision–language models, policy models, and cross-modal alignment techniques for manipulation, navigation, and planning, specifically focus on VLA architectures, organizing them into perception-aligned, policy-generative, and world-model-based categories, while identifying a unifying trend toward tightly integrated multi-modal interfaces \cite{xu2024surveyroboticsfoundationmodels}.

Large-scale datasets have been pivotal in enabling foundation-scale training. DROID~\cite{khazatsky2025droidlargescaleinthewildrobot} contributes over 150,000 trajectories spanning more than 1,000 objects and task scenarios, with multimodal annotations including RGB-D, language, low-dimensional states, and environment labels. The General Flow framework~\cite{yuan2024general} uses 3D point trajectories as transferable affordance representations, enabling cross-domain skill transfer from humans to robots. Similarly, ViSA-Flow~\cite{chen2025visa} pre-trains generative models on semantic action flows extracted from large-scale human–object interaction videos, requiring minimal adaptation for downstream robot learning.

Training strategies have also been extensively studied to improve efficiency and adaptability. Zhang et al.~\cite{zhang2024effectivetuningstrategiesgeneralist} analyze fine-tuning factors—including action space, policy head design, and supervision signals—through 2,500 rollout experiments, offering practical guidelines for adapting foundation-scale VLA models. Chen et al.~\cite{chen2025training} investigate the integration of chain-of-thought reasoning into embodied policy learning, demonstrating that lightweight reasoning mechanisms can yield significant performance gains with a 3× inference speedup compared to standard approaches.

Together, these efforts indicate a converging trajectory toward generalist embodied agents trained on massive, diverse datasets and equipped with modular reasoning capabilities. The combination of large-scale pre-training, efficient adaptation, and transferable affordance representations is positioning foundation-scale VLA models as the backbone for next-generation robotic intelligence. The representative works in this direction are summarized in Table~\ref{table:hybrid_architectures}(D), highlighting both data-centric and algorithmic advances driving foundation-scale VLA research.

\begin{figure*}
    \centering
    \includegraphics[width=1\linewidth]{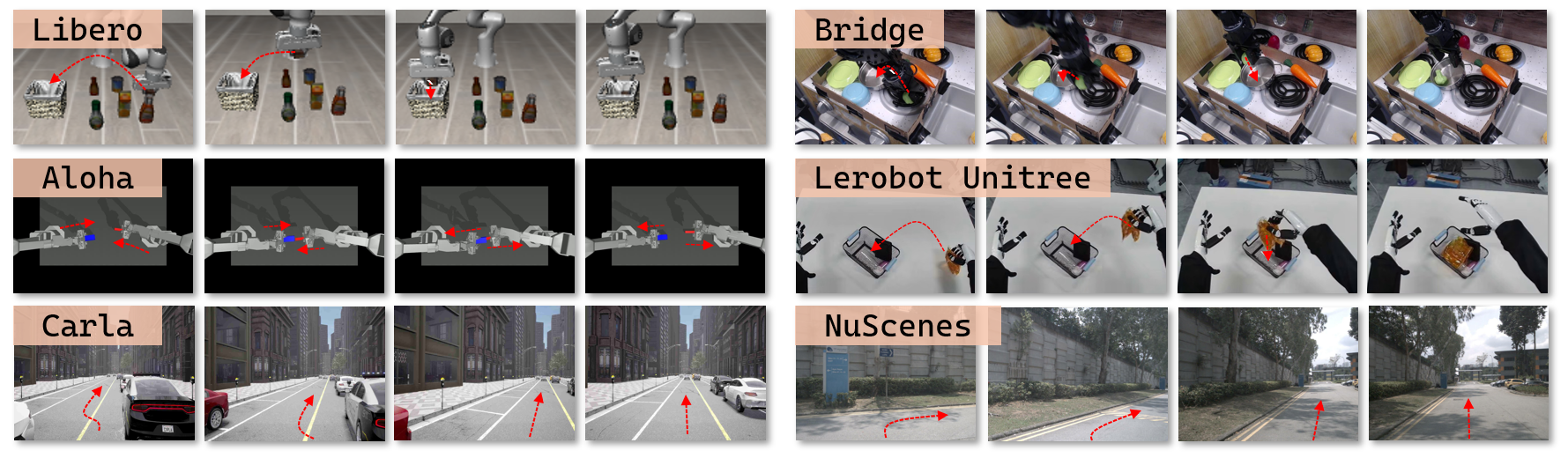}
    \vspace{-0.7cm}
    \caption{ Sample data of various datasets.}
    \vspace{-0.7cm}
    \label{fig:vla_4}
\end{figure*}

\subsubsection{Practical Deployment over Efficiency, Safety, and Human–Robot Collaboration}

As VLA models transition from research to real-world applications, practical deployment demands a holistic focus on efficiency, robustness, and human–robot interaction. Real-time inference, resilience to adversarial conditions, and seamless collaborative workflows are critical for reliable operation in dynamic, unpredictable environments. This direction integrates system optimization with safety and adaptability, ensuring that high-capacity models remain both effective and trustworthy in practice.

Efficiency-Oriented Designs have focused on reducing inference latency, lowering computational demands, and improving adaptability to resource-constrained platforms. For real-time execution, RTC (Real-Time Chunking)~\cite{black2025realtimeexecutionactionchunking} predicts upcoming action segments while executing current ones, enabling continuous high-frequency control. EdgeVLA~\cite{kscale2024evla} eliminates autoregressive dependencies in end-effector prediction and incorporates compact language models, achieving a 6× speedup with minimal performance degradation. Similarly, DeeR-VLA~\cite{yue2024deervladynamicinferencemultimodal} employs dynamic early-exit mechanisms to terminate inference once confidence thresholds are met, reducing online control costs.

Maintaining knowledge integrity during adaptation has become another priority. Knowledge-insulating VLA models~\cite{driess2025knowledgeinsulatingvisionlanguageactionmodels} address semantic degradation when integrating specialized modules into pretrained VLMs, using insulation strategies to retain cross-task generalization. Consistency-based acceleration strategies, such as CEED-VLA~\cite{song2025ceedvlaconsistencyvisionlanguageactionmodel}, apply consistency distillation and early-exit decoding to achieve over 4× inference acceleration while mitigating error accumulation through mixed-label supervision. Lightweight multi-modal fusion approaches like RoboMamba~\cite{liu2024robomambaefficientvisionlanguageactionmodel} and cross-domain adaptation methods such as ReVLA\cite{dey2025revlarevertingvisualdomain} further contribute to deployable efficiency.

Safety and Robustness have emerged as equally critical pillars for deployment readiness. SAFE~\cite{gu2025safe} leverages internal VLA feature representations to detect failures across multiple tasks, generalizing to unseen scenarios and enabling proactive intervention. Security assessments by Cheng et al.~\cite{cheng2024manipulationfacingthreatsevaluating} via the Physical Vulnerability Evaluation Procedures (PVEP) reveal vulnerabilities to adversarial patches, typography-based prompts, and distributional shifts, motivating the development of adversarially robust perception–control pipelines. Interpretability-focused work, such as Lu et al.~\cite{lu2025probingvisionlanguageactionmodelsymbolic}, uncovers symbolic encodings of objects, relations, and actions in VLA hidden layers, laying the groundwork for more transparent decision-making. Adaptive control frameworks like DyWA~\cite{lyu2025dywadynamicsadaptiveworldaction} further enhance robustness by jointly modeling geometry, state, physics, and action to respond to dynamic, partially observable conditions.

Human–Robot Collaboration research has explored interactive learning loops where humans and VLA models refine each other’s performance. Xiang et al.~\cite{xiang2025vla} propose collaborative frameworks that integrate limited expert interventions into VLA decision-making, reducing operator workload while enriching model training data. Closed-loop strategies like those in Zhi et al.~\cite{zhi2025closedloopopenvocabularymobilemanipulation} combine GPT-4V perception with real-time feedback control to adapt to environmental changes on the fly. History-aware policy learning~\cite{guhur2022instructiondrivenhistoryawarepoliciesrobotic} and object-centric visual prompting approaches such as CrayonRobo~\cite{li2025crayonroboobjectcentricpromptdrivenvisionlanguageaction} enhance task grounding and transparency, while skill library construction~\cite{li2025atomicskilllibraryconstruction} and grounding mask methods~\cite{huang2025roboground} enable scalable, reusable task decomposition. Camera-space policy designs like cVLA~\cite{argus2025cvla} improve sim-to-real transfer by predicting trajectory waypoints directly in 2D image coordinates, making policies more embodiment-agnostic. The representative methods for practical deployment are summarized in Table~\ref{table:hybrid_architectures}(E), highlighting key innovations across efficiency, safety, and human–robot collaboration.

In summary, practical deployment of VLA systems demands a multi-faceted design philosophy that simultaneously addresses efficiency, safety, and collaborative adaptability. The integration of real-time inference optimization, robustness against failures and adversarial conditions, and human-in-the-loop refinement strategies is paving the way for persistent, reliable, and interactive robotic systems in real-world environments.

\subsubsection{Discussion}
\textbf{Innovations}
The surveyed other advanced VLA highlight several innovations that collectively extend VLA research beyond the confines of previous section. Hybrid architectures that combine complementary paradigms for both reasoning and action generation, advanced multi-modal fusion for 3D-aware spatial grounding, and domain adaptations that extend VLA principles to areas such as autonomous driving, humanoid control, and GUI interaction. Foundation-scale models leverage massive multimodal datasets to build increasingly generalist agents, while deployment-oriented methods emphasize efficiency, safety, and human–robot collaboration for real-world applicability. 

\textbf{Limitations}
However, these hybrid systems remain computationally costly and complex to scale, and multi-modal fusion still struggles with noisy or incomplete real-world inputs. Domain-specific adaptations risk overfitting to narrow contexts, while foundation models demand prohibitive data and resource investments. Deployment efforts, though promising, continue to face challenges in robustness, interpretability, and reliability under adversarial or dynamic conditions. Addressing these limitations will require more efficient training strategies, broader evaluation standards, and tighter integration between research design and practical deployment.

\section{Datasets and Benchmarks} \label{sec:data}

\begin{table}[h]
\centering
\caption{Representative Datasets and Benchmarks for Robots.}
\label{realdatasettable}
\footnotesize
\begin{tabular}{p{2.8cm}p{0.5cm}p{1.2cm}p{1.2cm}p{0.8cm}}
\toprule
\textbf{Dataset Name} & \textbf{Year} & \textbf{Sensors} &  \textbf{Episodes}& \textbf{Tasks}\\
\midrule
\multicolumn{3}{l}{\textbf{(A) Real-world Datasets and Benchmarks}} \\
\midrule
MIME\cite{2c4cd2abfca5472f89be7a7df7884033} & 2018	&RGBD  &8300& 20 \\
EPIC-KITCHENS \cite{damen2018scalingegocentricvisionepickitchens} &2018 &RGB & 10000 &32 \\

RoboNet\cite{dasari2019robonet}	& 2019	&RGB & 162000& -- \\

MT-Opt\cite{kalashnikov2021mt}	& 2021	&RGB & 800000& 12 \\

BridgeData\cite{ebert2021bridge}	& 2021	&RGBD & 60100& 24 \\

Bc-z\cite{jang2022bc}	& 2022	&RGB & 25877 & 100 \\

RT-1\cite{brohan2022rt}	& 2022	&RGB& 13000 & 700 \\

MOO\cite{stone2023open}	& 2023	&RGB & 59100 & -- \\

RoboHive\cite{kumar2023robohive}	& 2023	&RGBD & 98500 & 38 \\
OBRH \cite{shafiullah2023bringingrobotshome} &2023 &RGBD & 5600 &109 \\

RH20T\cite{fang2024rh20t}	& 2024	&RGBD & 110000 & 147 \\

DROID\cite{khazatsky2024droid}	& 2024	&RGBD & 76000 & -- \\

AutoRT\cite{ahn2024autort}	& 2024	&RGB & 77000 & -- \\

UMI\cite{chi2024universal}	& 2024	&RGB & 1400 & 7 \\

OXE\cite{o2024open}	& 2025	&RGBD & \textgreater 1000000 & 160266 \\

\midrule
\multicolumn{3}{l}{\textbf{(B) Simulation Datasets and Benchmarks}} \\
\midrule

ROBOTURK \cite{pmlr-v87-mandlekar18a}	& 2018 &RGB  & \textgreater7868& 2200 \\

Meta-World \cite{yu2020meta} &2019	&RGB &--& 50 \\


ALFRED \cite{shridhar2020alfred}& 2019	&RGBD  &8055& 120 \\

RLBench \cite{james2020rlbench} & 2019	&RGBD &--& 100\\


VIMA 0.5 \cite{jiang2022vima}& 2022	&RGB  & \textgreater 600000& \textgreater 1000 \\

CALVIN 0.5 \cite{mees2022calvin}& 2022	&RGBD  &--& 34 \\

LIBERO \cite{liu2023liberobenchmarkingknowledgetransfer} &2023 &RGB & 5000 &100 \\
Lota-Bench \cite{choi2024lotabenchbenchmarkinglanguageorientedtask}	& 2024 &RGB&611& 12 \\
Mobile ALOHA \cite{fu2024mobilealohalearningbimanual} &2024 &RGB& 825 &7 \\
RoboCasa \cite{nasiriany2024robocasalargescalesimulationeveryday} &2024 &RGBD & \textgreater 100000 &100 \\
RoboGen \cite{wang2024robogen}	& 2024 &RGB  &--& 106 \\

\bottomrule
\end{tabular}
\end{table}

Like other imitation learning approaches, Visual Language Action (VLA) models rely on high-quality labeled datasets. These datasets are either collected from real-world scenarios or generated using simulation environments, the dataset samples are illustrated in Fig. \ref{fig:vla_4}. Typically, they contain multimodal observations—such as images, LiDAR point clouds, and inertial measurement unit (IMU) readings—along with corresponding ground-truth labels and language instructions. To facilitate a systematic understanding, we analyze existing datasets and benchmarks, and propose a taxonomy that organizes datasets according to their complexity, modality, and task diversity. This taxonomy provides a clear framework for evaluating the suitability of different datasets for VLA research and highlights potential gaps in existing resources, the representative works are summarized in Table~\ref{realdatasettable}.

\subsection{Real-World Datasets and Benchmarks} 

High-quality real-world datasets are fundamental for the development of reliable VLA algorithms. In recent years, numerous high-quality and diverse real-world robotics datasets have been collected. Researchers have gathered datasets using different sensor modalities, across a variety of tasks and environmental settings.

\subsubsection{Real-World Datasets and Benchmarks for Embodied Robotics} 

A real-world embodied robotics dataset refers to a collection of multimodal data acquired from robots that interact with their environments through perception and action. 
Embodied robotics datasets are specifically designed to capture the complex interactions between visual, auditory, proprioceptive, and tactile sensory inputs and the corresponding motor actions, intentions, and environmental contexts.
They are essential for training and evaluating models in embodied AI, where the objective is to enable robots to perform tasks through closed-loop, adaptive behaviors in dynamic environments. 
By providing rich, temporally aligned observations and actions, these datasets serve as foundational resources for developing and benchmarking algorithms in imitation learning, reinforcement learning, visual-language action, and robotic planning.

Current embodied robot datasets faces significant data costs issues because real-world robot data is not largely collected. Collecting real-world robot datasets poses many challenges. It not only requires hardware equipments, but also requires precise manipulation. Among them, MIME \cite{2c4cd2abfca5472f89be7a7df7884033}, RoboNet \cite{dasari2019robonet}, and MT-Opt \cite{kalashnikov2021mt} have collected large-scale robot demonstration datasets spanning a range of tasks, from simple object pushing to complex household object stacking. Different from prior datasets that often assume a single optimal trajectory per task, these datasets include multiple demonstrations for the same task, using the minimum distance among test trajectories as an evaluation metric. This approach has significantly advanced research in manipulation and VLA tasks.
BridgeData \cite{ebert2021bridge} offers a large-scale, multi-domain robot dataset comprising 71 tasks across 10 environments. Experiments demonstrate that jointly training on this dataset along with a small subset (e.g., 50 tasks) of unseen tasks in a new domain can double success rates compared to using target domain data alone. As a result, many contemporary VLA methods adopt BridgeData for model training.
In the embodied AI field, model generalization is often limited by the difficulty of collecting diverse real-world robotic data. RT-1 \cite{brohan2022rt} provides a broad dataset of real-world robotic tasks to improve both task performance and generalization to novel scenarios. Similarly, Bc-z \cite{jang2022bc} includes previously unseen manipulation tasks involving novel combinations of objects within the same scene, supporting research in generalizable policy learning.
Several datasets also provide comprehensive software platforms and ecosystems for Embodied AI, covering environments such as hand manipulation, locomotion, multi-tasking, multi-agent scenarios, and muscle-based control \cite{stone2023open}. Compared to earlier works, RoboHive \cite{kumar2023robohive} bridges the gap between current robot learning capabilities and potential growth, supporting diverse learning paradigms including reinforcement, imitation, and transfer learning.
Distinctively, RH20T \cite{fang2024rh20t} offers 147 tasks encompassing 110K manipulation episodes, including multimodal visual, force, audio, and action data. Each episode is accompanied by a human demonstration and language description, making this dataset particularly suitable for one-shot imitation learning and policy transfer to new tasks based on previously trained episodes.

To advance the development of more generalizable manipulation policies, the robotics community must prioritize the collection of large-scale, diverse datasets spanning a wide range of tasks and environmental settings. Several datasets have been collaboratively gathered by multiple robots across different regions, making them among the most geographically and contextually diverse embodied robot datasets to date \cite{khazatsky2024droid, ahn2024autort, chi2024universal, guruprasad2024benchmarkingvisionlanguage}. In addition, Open X-Embodiment (OXE) \cite{o2024open} consolidates 22 robot datasets collected through collaboration among 21 institutions, covering 527 skills and 160,266 tasks. OXE provides standardized data formats to facilitate easy use by researchers. An overview of these datasets is provided in Table~\ref{realdatasettable} (A).

For benchmark evaluation, researchers typically use Success Rate—the proportion of tasks successfully completed relative to the total number of tasks. Some studies additionally employ Language Following Rate to assess the model’s ability to interpret and execute language instructions. Furthermore, recent VLA models are often evaluated by transferring trained policies to previously unseen environments to measure robustness and generalization performance \cite{black2024pi0visionlanguageactionflowmodel, kim2024openvlaopensourcevisionlanguageactionmodel}.

\subsubsection{Real-World Datasets and Benchmarks for Autonomous Driving} 

Autonomous driving dataset is different from embodied robot dataset. It has emerged as one of the most transformative applications of artificial intelligence, relying heavily on large-scale datasets to train and evaluate perception, planning, and control algorithms. High-quality datasets are foundational for the development of robust and generalizable autonomous driving systems, as they enable supervised learning, benchmarking, and simulation of rare or safety-critical scenarios. Over the past decade, numerous datasets such as, \cite{openlane, cityscapes, bdd, apollo, argoverse, navsim, once, waymo, nuscenes, kitti}, have been introduced, offering multi-modal sensor data including camera images, LiDAR point clouds, radar signals, and high-definition maps. These datasets vary significantly in geographic coverage, sensor configuration, driving behavior diversity, and annotation richness, making them complementary resources for research and development. 

However, most public datasets are collected in open-loop settings and primarily represent normal driving behavior, which limits their ability to cover long-tail corner cases. To address this gap, recent efforts have focused on generating synthetic data, simulating closed-loop interactions, and curating datasets tailored to rare or safety-critical events. Continued innovation in dataset design remains vital for advancing safe, scalable, and generalizable autonomous driving systems. 

For evaluation, autonomous driving VLA models commonly rely on metrics such as L2 distance, which measures deviation from reference trajectories—and completion rate, which quantifies the proportion of successfully completed driving tasks.

\subsection{Simulation Datasets and Benchmarks}
Collecting large-scale real-world data for continuous control tasks poses significant challenges, as these tasks require real-time interaction and continuous feedback from human annotators. Moreover, acquiring such data is often costly and time-consuming, limiting its scalability.
This enables a scalable mechanism for diverse human supervision on an extensive set of problem instances. To investigate the performance of embodied robot or autonomous driving models in large-scale and high-quality data, researchers utilize the simulated data from virtualization engines for training and evaluation.

\subsubsection{Simulation Datasets and Benchmarks for Embodied Robotics}

Simulation datasets for embodied AI typically include synthetic scenes, physics-based interactions, annotations for navigation, object manipulation, task execution, and agent-environment dynamics. These datasets allow for benchmarking and training across a wide range of tasks, from visual navigation and semantic exploration to complex multi-step object manipulation. Prominent examples include \cite{puig2018virtualhome,yu2020meta,gupta2020relay,james2020rlbench,wang2024robogen}, each offering different trade-offs in realism, task diversity, and control fidelity.
By enabling safe experimentation and massive-scale data collection, simulation datasets are foundational to the development of robust, generalizable embodied agents. As the field matures, the design of richer, more realistic simulation datasets, covering diverse embodiments, tasks, and environments, continues to drive progress toward real-world deployment.

ROBOTURK \cite{pmlr-v87-mandlekar18a} is a simulated dataset for high quality 6-DoF manipulation states and actions, collected through teleoperation using mobile devices. Unlike traditional approaches that rely on remote users to demonstrate actions within a virtual engine, ROBOTURK leverages policy learning to generate multi-step robot tasks with varying rewards. By aggregating large quantities of demonstrations, the dataset provides precise and reliable data for both training and evaluation.
iGibson0.5 \cite{xia2020interactive} introduces a benchmark for training and evaluating interactive navigation solutions. This work not only provides a novel experimental simulation environment but also proposes a dedicated metric to assess the interplay between navigation and physical interaction along navigation paths. The benchmark introduces the Interactive Navigation Score, composed of two sub-metrics: Path Efficiency and Effort Efficiency. Path Efficiency is defined as the ratio between the length of the shortest successful path and the actual path length traversed by the robot, weighted by a success indicator function. Effort Efficiency captures the excess kinematic and dynamic effort required during navigation, reflecting the cost of physical interactions.
VIMA \cite{jiang2022vima} introduces a new benchmark, VIMA-BENCH, which establishes a four-level evaluation protocol to assess progressively stronger generalization capabilities, ranging from randomized object placement to entirely novel tasks. Similarly, CALVIN and LOTA-Bench \cite{mees2022calvin, choi2024lotabenchbenchmarkinglanguageorientedtask} focus on learning long-horizon, language-conditioned tasks across diverse manipulation environments using multimodal robot sensor data. These benchmarks are particularly suited for evaluating methods that aim to generalize to unseen entities by training on large-scale interaction datasets and testing on novel scenes. Performance in these benchmarks is typically measured using task success rates.
An overview of these simulation datasets is provided in Table~\ref{realdatasettable}(B).

\subsubsection{Simulation Datasets and Benchmarks for Autonomous Driving}

Closed-loop simulation plays a critical role in ensuring the safety of autonomous driving systems, as it enables the generation of safety-critical scenarios that are difficult or dangerous to capture in the real world. 
While previously recorded driving logs provide valuable resources for constructing new scenarios, closed-loop evaluation demands modifications to the original sensor data in order to reflect updated scene configurations. For example, actors may need to be added or removed, and the trajectories of both existing actors and the ego vehicle may differ from those in the original recordings \cite{bench2drive, lmdrive}.
UniSim \cite{Yang_2023_CVPR} is a neural sensor simulator that extends single recorded trajectories into multi-sensor closed-loop simulations. It constructs neural feature grids to reconstruct both static backgrounds and dynamic actors, compositing them to simulate LiDAR and camera data from novel viewpoints. This allows for the addition, removal, or repositioning of actors. To better accommodate unseen viewpoints, UniSim further employs a convolutional network to complete regions not visible in the original data.

Unlike real-world autonomous driving datasets, closed-loop simulation benchmarks require specialized evaluation metrics tailored to interactive driving tasks. Commonly used metrics include Driving Route (measuring adherence to planned trajectories), Infraction Score (penalizing traffic rule violations), and Completion Score (assessing task completion). Together, these metrics provide a more comprehensive assessment of VLA model performance in realistic, safety-critical driving scenarios.

\subsection{Discussion}


\textbf{Innovations}
This paper introduce systematic taxonomies, standardized evaluation metrics, and large-scale collaborative efforts such as Open X-Embodiment (OXE), which unifies datasets from multiple institutions to promote reproducibility and generalization. These contributions enable broader coverage of tasks, richer modality combinations, and improved policy transfer across domains, advancing the scalability of embodied AI research. 

\textbf{Limitations}
However, real-world datasets are expensive and logistically challenging to collect, often restricted to controlled laboratory environments with limited scene diversity; simulation datasets, while scalable and safe, still struggle to fully capture the complexity, noise, and unpredictability of real-world interactions. Moreover, benchmark metrics such as success rate and trajectory deviation may insufficiently reflect nuanced capabilities like language grounding, long-horizon reasoning, or safe deployment in unstructured environments. Addressing these limitations requires not only expanding dataset diversity and realism but also designing richer evaluation protocols that better capture the demands of real-world autonomy.

\section{Simulators}\label{sec:sim}

Robot simulators have become indispensable tools for developing and evaluating intelligent robotic systems in diverse and interactive environments. These platforms typically integrate physics engines, sensor models (e.g., RGB-D, IMU, LiDAR), and task logic to support a wide range of tasks such as navigation, manipulation, and multimodal instruction following. State-of-the-art simulators provide scalable, photorealistic, and physically plausible environments for training embodied agents using reinforcement learning, imitation learning, or large pre-trained models. By offering safe, controllable, and reproducible settings, embodied simulators accelerate the development of generalizable robotic intelligence while significantly reducing the cost and risks associated with real-world experimentation \cite{puig2018virtualhome, tassa2018deepmind}.

THOR \cite{kolve2017ai2} is a simulator featuring near photo-realistic 3D indoor scenes, where AI agents can navigate environments and interact with objects to complete tasks. It supports a variety of research areas, including imitation learning, reinforcement learning, manipulation planning, visual question answering, unsupervised representation learning, object detection, and semantic segmentation. In contrast, some simulators are based on virtualized real spaces rather than artificially designed environments, encompassing thousands of full-scale buildings equipped with embodied agents that are subject to realistic physical and spatial constraints \cite{xia2018gibson, xiang2020sapien}. Habitat \cite{savva2019habitat} and Habitat 2.0 \cite{szot2021habitat} further extend this paradigm by offering scalable simulation platforms for training embodied agents in complex 3D environments with interactive, physics-enabled scenarios.
ALFRED \cite{shridhar2020alfred} introduces a benchmark consisting of long-horizon, compositional tasks with non-reversible state changes, aiming to bridge the gap between simulation benchmarks and real-world applications. ALFRED includes both high-level goals and low-level language instructions, making tasks significantly more complex in terms of sequence length, action space, and linguistic variability compared to existing vision-and-language datasets.

Earlier simulation environments that combined physics and robotic tasks often focused on a narrow set of scenarios and featured only small-scale, simplified scenes. In contrast, iGibson 1.0 \cite{shen2021igibson} and iGibson 2.0 \cite{li2022igibson} are open-source simulation platforms that support a more diverse set of household tasks within large-scale, realistic environments. Their scenes are replicas of real-world homes, with object distributions and layouts closely aligned with physical spaces, thereby enhancing ecological validity and bridging the gap between simulated and real-world robotic learning.

Advanced simulators not only allow multiple agents to interact within the same environment but also provide a wide range of sensor and physics outputs. Ideally, such simulators should combine a universal physics engine, a flexible robotics simulation platform, and a high-fidelity rendering system. These features make them powerful tools for both robotic simulation and generative model evaluation \cite{gan2021threedworld,10027470, bullet3, authorsgenesis, bu2025agibot, tao2025maniskill3gpuparallelizedrobotics}.

MuJoCo \cite{todorov2012mujoco} is a widely adopted, open-source physics engine designed to facilitate research and development in robotics and related domains that require accurate simulation.
More recently, GPU-based simulation engines have gained popularity. Notably, NVIDIA Isaac Gym \cite{makoviychuk2021isaacgymhighperformance}, built on the Omniverse platform, enables large-scale development, simulation, and testing of AI-driven robots in physically realistic virtual environments. Isaac Gym has become increasingly popular in both academia and industry for accelerating the creation of new robotics tools and enhancing existing systems.

Similar challenges exist in autonomous driving, where large-scale real-world data collection and annotation are both costly and time-consuming. Collecting sufficient data to cover the multitude of rare corner cases is particularly difficult. To address this, researchers have developed simulators that include both static road elements (e.g., intersections, traffic lights, and buildings) and dynamic agents (e.g., vehicles and pedestrians). CARLA \cite{carla} and LGSVL \cite{lgsvl} leverage game engines to render realistic driving scenarios, supporting flexible sensor configurations and generating signals suitable for training and evaluating driving strategies. These platforms have become critical for advancing autonomous driving research by providing controllable, reproducible, and cost-effective testing environments.

\section{Robot Hardware}\label{sec:hardw}
The physical structure of a robot provides the foundation for perception, locomotion, manipulation, and interaction with its environment. Its core components typically include sensors, actuators, power systems, and a control unit. Sensors—such as cameras, LiDAR, inertial measurement units, and tactile arrays—supply essential information about both the external environment and the robot’s internal state. Actuators, including motors, servos, or hydraulic systems, convert control signals into physical actions, enabling tasks such as locomotion and object manipulation. The control unit, generally based on embedded processors or microcontrollers, functions as the computational core by integrating sensor inputs and issuing commands to actuators. Power systems, typically in the form of batteries or external energy sources, sustain continuous operation. Hardware design must balance performance, energy efficiency, weight, and durability to satisfy task-specific requirements across diverse application domains, including industrial automation, service robotics, and autonomous vehicles \cite{franka, galaxea, lerobot}.

\section{Challenges and Future Directions}\label{sec:chall}

\subsection{Challenges of Vision-Language-Action Models}

This section summarizes the open challenges and future directions in advancing Vision-Language-Action (VLA) models. Despite remarkable progress in recent years, the development of VLA models has gradually revealed critical bottlenecks. The most fundamental issue lies in the fact that current VLA systems are largely built upon the transfer of large-scale LLMs or VLMs. Although these models excel in semantic understanding and cross-modal alignment, they lack direct training and experience in interacting with the physical world. As a result, VLA systems often demonstrate the gap of “understanding the instruction but failing to execute the task” in real environments. This reflects a fundamental contradiction: the disconnection between semantic-level generalization and embodied capabilities in the physical world. How to achieve the transformation from non-embodied knowledge to embodied intelligence, and to truly bridge the gap between semantic reasoning and physical execution, remains the central challenge. Specifically, this contradiction is manifested in the following aspects.

\subsubsection{Scarcity of Robotic Data} 
Robotic interaction data is a critical resource that determines the performance of VLA models; however, existing datasets remain insufficient in both scale and diversity. Collecting large-scale demonstrations across a wide range of tasks and environments in the real world is constrained by hardware costs, experimental efficiency, and safety concerns. Existing open-source datasets, such as 
Open X-Embodiment, have advanced robot learning but are primarily focused on tabletop manipulation and object grasping. This lack of task and environmental diversity significantly limits generalization to novel contexts and complex tasks. Simulation platforms, such as RLBench, offer a cost-effective means of generating large-scale trajectories but are constrained by rendering fidelity, physics engine accuracy, and task modeling limitations. Even with techniques such as domain randomization or style transfer, the sim-to-real gap persists, with many models performing well in simulation yet failing when deployed on physical robots. Therefore, enhancing both the diversity and realism of robotic data at scale remains a primary challenge in mitigating generalization deficits.

\subsubsection{Architectural Heterogeneity} 
Most VLA models attempt end-to-end modeling across vision, language, and action, but their implementations reveal strong heterogeneity. On one hand, different works adopt different backbone networks: vision encoders may rely on ViT, DINOv2, or SigLIP; language backbones on PaLM, LLaMA, or Qwen; and action heads on discrete tokenization, continuous control vectors, or even diffusion-based generation. Such architectural diversity hinders comparison and reuse across models, slowing the emergence of unified standards. On the other hand, perception, reasoning, and control are often loosely coupled internally, leading to fragmented feature spaces and weak portability across platforms or task domains. Some models excel at cross-task language understanding but require heavy adaptation when interfacing with low-level controllers. This architectural heterogeneity increases integration complexity and significantly constrains generalization and scalability.

\subsubsection{Real-Time Inference Constraints and Cost} 
Current VLA models are heavily dependent on large-scale Transformer architectures with autoregressive decoding, which severely limits inference speed and execution efficiency on real robots. Since each action token depends on the previous one, latency accumulates, while high-frequency tasks such as dynamic grasping or mobile navigation demand millisecond-level responses. Moreover, high-dimensional visual inputs and massive parameter counts impose prohibitive computation and memory costs. Many state-of-the-art VLAs require GPU memory far beyond the capacity of typical embedded platforms. Even with quantization, compression, or edge–cloud co-inference, it remains difficult to balance accuracy, real-time performance, and low cost. This combination of inference constraints and hardware bottlenecks leaves VLA deployment trapped between being too slow and too expensive.

\subsubsection{Pseudo-Interaction in Human–Robot Interaction} 
Systems generate actions based on prior knowledge or static training patterns rather than engaging in genuine interaction grounded in environmental dynamics and causal reasoning. When encountering unfamiliar settings or state changes, models often rely on statistical co-occurrence learned from data, rather than probing the environment or using sensor feedback to refine actions. This lack of causal reasoning means that VLAs may appear to follow instructions but fail to establish genuine causal chains between environment states and action outcomes. 
As a result, robots often fail to adapt in dynamic environments. 
This pseudo-interaction highlights VLA’s deficiency in causal modeling and feedback utilization, and remains a key obstacle to embodied intelligence.

\subsubsection{Evaluation and Benchmarking Limitations} 
The evaluation of VLA models is also limited. Current benchmarks are mainly set in laboratory or highly structured simulated environments, focusing on tabletop manipulation or object grasping. While such tasks measure performance on narrow distributions, they do not capture generalization or robustness in open-world scenarios. Once deployed in outdoor, industrial, or complex home settings, performance often degrades drastically, exposing the gap between evaluation and real-world applicability. 
This narrow evaluation scope hinders comprehensive assessment of VLA feasibility and limits horizontal comparisons across models. The lack of unified, authoritative, and diverse benchmarks is becoming a major bottleneck for real-world progress.

While these five aspects highlight key shortcomings in data, architecture, interaction, and evaluation, they do not exhaust the challenges faced by VLA research. More fundamentally, the long-term question is whether VLA systems can truly achieve controllability, trustworthiness, and safety. In other words, the future of VLA requires not only solving performance and generalization issues but also addressing the deeper concerns of deploying intelligent agents responsibly. This transition implies that researchers must move beyond model optimization toward systemic paradigm shifts in order to meet long-term challenges.




\subsection{Opportunities of Vision-Language-Action Models}

Despite formidable challenges, the future of VLA is also full of opportunities. As the crucial bridge connecting language, perception, and action, VLA has the potential to transcend the semantic–physical gap and become a central pathway for embodied intelligence. Overcoming current bottlenecks could reshape the paradigm of robotics research and position VLA at the forefront of real-world deployment.

\subsubsection{World Modeling and Cross-Modal Unification} 
At present, language, vision, and action remain loosely coupled in VLA systems, confining them to instruction “generation” rather than holistic world understanding. Achieving true cross-modal unification would enable VLA to model environments, reasoning, and interaction jointly within a single token stream. This unified structure would allow VLA to evolve into a proto–world model, enabling robots to close the loop from semantic understanding to physical execution. Beyond a technical advance, this would mark a crucial step toward artificial general intelligence.

\subsubsection{Breakthroughs in Causal Reasoning and Genuine Interaction} 
Most existing VLAs rely on static data distributions and surface-level correlations, lacking the ability to interact based on causal laws. They simulate interaction by guessing from prior patterns, rather than probing the environment and updating strategies with feedback. If future VLAs can incorporate causal modeling and interactive reasoning, robots will learn to probe, validate, and adapt—enabling genuine dialogue with dynamic environments. Such a breakthrough would overcome pseudo-interaction and mark the transition from data-driven intelligence to deeply interactive intelligence.

\subsubsection{Virtual–Real Integration and Large-Scale Data Generation} 
While data scarcity is a critical limitation, it also represents a massive opportunity. If virtual and real data ecosystems can be integrated—through high-fidelity simulation, synthetic data generation, and multi-robot sharing—it will be possible to construct datasets containing trillions of trajectories across diverse tasks. Just as GPT leveraged internet-scale corpora to trigger a leap in language intelligence, such data ecosystems could trigger a leap in embodied generalization, enabling VLAs to operate robustly in open-world scenarios.

\subsubsection{Societal Embedding and Trustworthy Ecosystems} 
The ultimate value of VLA lies not only in technical capability but also in societal integration. As VLA enters public and domestic spaces, safety, trustworthiness, and ethical alignment will determine its adoption. Establishing standardized frameworks for risk assessment, explainability, and accountability would transform VLAs from laboratory artifacts into trusted partners. Once embedded in society, VLAs could serve as the next-generation human–AI interface, reshaping domains such as healthcare, industry, education, and services. This societal embedding marks a milestone opportunity for translating frontier research into real-world transformation.

\section{Conclusion}

Recent advances in Vision-Language-Action (VLA) models have extended the generalizable capabilities of vision-language models to robotic applications, including embodied AI, autonomous driving, and diverse manipulation tasks. This survey systematically charts the emergence of VLA approaches by examining their motivations, methodologies, and applications. It provides a unified taxonomy of architectural structures and analyzes over 300 articles, along with supporting materials. We first categorize VLA architectural innovations based on autoregression-based models, diffusion-based models, reinforcement-based learning, hybrid structures, and efficiency optimization techniques. Subsequently, we explore the datasets, benchmarks, and simulation platforms that facilitate VLA training and evaluation. Building on this comprehensive review, we analyze the strengths and limitations of current approaches and highlight potential directions for future research. Collectively, these insights offer a consolidated reference and a forward-looking roadmap for developing trustworthy, continually evolving VLAs capable of advancing general artificial intelligence in robotic systems.


	
\tiny
\bibliographystyle{IEEEtran}
\bibliography{references}


	
\end{document}